# Open-access model for detecting openly dumped dispersed municipal solid waste from crowdsourced UAV imagery in Sub-Saharan Africa

Steffen Knoblauch [a,b,c,*], Ram Kumar Muthusamy [a,c], Luís M.A. Bettencourt [d,e,f], Costas Velis [g], Pierre Chrzanowski [h], Edward Charles Anderson [h], Pete Masters [i], Innocent Maholi [j], Antonio Inguane [k], Levi Szamek [a], Alexander Zipf [a,b,c]

[a] HeiGIT at Heidelberg University, Heidelberg, Germany
[b] Interdisciplinary Centre of Scientific Computing (IWR), Heidelberg University, Heidelberg, Germany
[c] GIScience Research Group, Heidelberg University, Heidelberg, Germany
[d] Urban Science Laboratory, Department of Ecology and Evolution, The University of Chicago, Chicago, IL, USA
[e] Santa Fe Institute, Santa Fe, NM, USA
[f] Complexity Science Hub, Vienna, Austria
[g] Department of Civil and Environmental Engineering, Imperial College London, London, United Kingdom
[h] Global Facility for Disaster Reduction and Recovery (GFDRR), World Bank, Paris, France
[i] Humanitarian OpenStreetMap Team, United Kingdom
[j] Open Map Development Tanzania, Dar Es Salaam, Tanzania
[k] Data4Moz, Chimoio, Mozambique

## ARTICLE INFO

## ABSTRACT

Managing municipal solid waste in rapidly urbanizing Sub-Saharan Africa remains challenging due to dispersed informal dumping and limited high-resolution datasets for spatial monitoring. We present an open-access deep learning model for automated detection of openly dumped dispersed MSW (ODD-MSW) via crowdsourced UAV imagery, trained and evaluated across 29 regions in 10 Sub-Saharan African countries, encompassing diverse environmental contexts. A deep learning model trained on manually annotated image tiles achieved an overall accuracy of 92.87% (F1 = 0.927) in detecting ODD-MSW across all study regions. Applied without any site-specific retraining to 25 independent OpenAerialMap scenes from 21 cities in 19 countries on five continents, the frozen model retained an F1 score of 0.867, providing direct evidence of transferability beyond the training domain. Predicted distributions reveal heterogeneous accumulation patterns, ranging from localized hotspots — often along waterways, where waste can exacerbate flood and public health risks — to more dispersed litter across urban areas. Waste accumulation is most strongly associated with population density (Spearman $\rho = 0.674$, $p < 0.001$) and indicators of lack of local infrastructure access ($\rho = 0.700$, $p < 0.001$), whereas its relationship with broader measures of regional development is weaker and not statistically significant ($\rho = 0.215$, $p = 0.26$), highlighting the importance of fine-scale data for understanding localized waste dynamics. By releasing the model, this study provides a ready-to-use tool for UAV imagery collected by municipalities and local mapping communities, enabling ODD-MSW monitoring without extensive technical expertise. This approach empowers local practitioners to convert UAV imagery into actionable insights, supporting targeted interventions and improved municipal solid waste management across Sub-Saharan Africa.

## 1. Introduction

Municipal solid waste (MSW) management is one of the defining environmental and public health challenges of the twenty-first century (United Nations, 2015). Global annual MSW production is projected to increase from approximately 2.0 to 3.4 billion tonnes by 2050 driven by rapid urbanization, economic growth and changing consumption patterns (Environment, 2024; Kaza et al., 2018). The problem is particularly acute in densely populated urban areas, where large volumes of MSW easily accumulate in confined and public spaces and infrastructure development frequently lags behind demand, leading to informal disposal (Cook et al., 2026). While MSW comprises both organic and inorganic materials and can represent a valuable secondary resource

* Corresponding author.
*E-mail address:* steffen.knoblauch@uni-heidelberg.de (S. Knoblauch).

Open access funding enabled and organized by Projekt DEAL.

when properly managed (Di Maria et al., 2013), its mismanagement has profound environmental and societal consequences. Open dumping contaminates freshwater systems (Lambert & Wagner, 2017), degrades soils, reduces agricultural productivity (Kihampa & Mwegoha, 2011; Rouhani & Hejcman, 2025) and impairs drainage networks, thereby exacerbating flood risk especially in areas with unplanned development and limited waste and storm water management systems (Carbery et al., 2018; Mokuolu et al., 2022; Moritz & Knoblauch, 2023). Accumulated waste also creates breeding grounds for disease vectors, such as *Aedes* mosquitoes, and thus underpins the prevalence and spread of vector-borne diseases (Knoblauch, Su Yin et al., 2024). Plastic debris further threatens marine ecosystems and enters food chains as microplastics with well documented adverse health consequences (Carbery et al., 2018; Cook et al., 2022; Gkoutselis et al., 2021; Leslie et al., 2022; Osman et al., 2023).

Global policy commitments have recognized the generality of these challenges and called for increases in institutional capacity as well as social and technological innovations to address them. This includes specific objectives set in the context of the United Nations Sustainable Development Goals (United Nations, 2015), particularly sustainable consumption and production (Goal 12) and sustainable urbanization (Goal 11). Goal 11.6 targets MSW collection coverage specifically (UN Habitat, 2021). This momentum is further reflected in transnational city networks such as C40 Cities, where 14 large cities across Africa, Asia, and Latin America have committed to the Sustainable Waste Systems Accelerator, setting targets and fostering collaborations to transform MSW management, reduce emissions, generate employment, and improve public health outcomes (C40 Cities, 2025a). As of 2025, participating cities report measurable progress, although the uptake of digital MSW management technologies remains uneven (C40 Cities, 2025b). Emerging on-demand platforms — such as DortiBox in Sierra Leone, Yo-Waste in Uganda, Baus Taka in Kenya, Coliba in Ghana, and the WRAP Project in South Africa — illustrate both the potential and fragmentation of digital innovation in this domain (Baus Taka, 2026; Coliba, 2026; DortiBox, 2026; WRAP, 2026; Yo-Waste, 2026). Despite this increase in attention to the issue, an estimated three billion people still lack access to controlled MSW disposal (Wilson & Velis, 2015), and illegal dumping has remained widespread—particularly in rapidly growing cities in Sub-Saharan Africa, where formal MSW collection often covers less than half of generated MSW (Gwada et al., 2019; Kaza et al., 2018; Lebreton & Andrady, 2019).

Sub-Saharan Africa is undergoing rapid demographic and urban transformation, with population growth, urban expansion and economic development driving a sharp increase in MSW generation (Combes et al., 2025). The region generated an estimated 231 million tonnes of MSW in 2022, accounting for approximately 9% of global production (Cook et al., 2026). Despite relatively low per capita generation rates (0.52 kg per person per day), MSW management systems exhibit the lowest coverage worldwide, averaging only 31% collection (Cook et al., 2026). Service provision is highly uneven, reaching around 45% of urban populations but only 6% in rural areas (Cook et al., 2026), leaving the majority of MSW unmanaged and frequently disposed of through open dumping or burning. Even collected MSW is predominantly directed to uncontrolled dumpsites, with minimal recycling or sanitary disposal. With MSW generation projected to more than double by 2050 — the fastest growth rate globally — these trends are expected to further intensify the pressure on already constrained systems. MSW management systems in the region are further challenged by fragmented governance structures, uneven service provision and chronic underfunding. Collection services are delivered through a mix of municipal authorities, private contractors, and informal or community-based providers, resulting in pronounced spatial inequalities. In the absence of effective regulatory oversight, lower-income and rapidly expanding urban areas are often underserved, leading to widespread informal dumping and uncollected MSW. Limited financial resources and weak institutional capacity constrain investments in infrastructure, maintenance and monitoring. As a result, reliable, spatially explicit data on open dumping remain scarce, hindering targeted interventions and evidence-based planning.

In this context, addressing dispersed open dumping requires spatially explicit and scalable monitoring systems capable of identifying not only large, formal dumpsites but also small and informal MSW accumulations embedded within urban fabrics (Fraternali et al., 2024; Kalonde et al., 2025). However, comprehensive official datasets are rare, inconsistent or outdated, and typically overlook openly dumped dispersed MSW (ODD-MSW) that nonetheless contributes substantially to environmental degradation and infrastructure failure (Lynch, 2018; Tisserant et al., 2017). Citizen science initiatives have helped narrow this information gap, yet their reliance on volunteer participation introduces spatial biases and limits temporal consistency and scalability (Haarr et al., 2020; Irwin, 1995; Jambeck & Johnsen, 2015; Lynch, 2018). Recent advances in computer vision and deep learning have demonstrated strong performance in automated MSW detection, particularly on curated benchmark datasets (Abdu & Mohd, 2022). When combined with satellite remote sensing, these approaches can identify large landfills or extensive MSW sites at regional scales (Devesa & Brust, 2021; Sun et al., 2023; Zeng et al., 2019). However, even very high-resolution satellite imagery often lacks the spatial detail required to detect small, ODD-MSW accumulations (Fraternali et al., 2024). Unmanned aerial vehicle (UAV) imagery provides the necessary centimeter-scale resolution. Though very recent, these new datasets have already demonstrated considerable promise for high-resolution mapping of ODD-MSW (Kalonde et al., 2025; Knoblauch et al., 2025; Knoblauch, Szamek et al., 2024; Morandini et al., 2026; Pan et al., 2022; Papale et al., 2023; Sliusar et al., 2022; Wang et al., 2024; Wyard et al., 2021). While some studies have employed semantic segmentation to achieve pixel-level delineation of MSW, enabling precise spatial boundaries of waste objects, such approaches typically require extensive pixel-wise annotation and may be sensitive to domain shifts across heterogeneous UAV datasets. In contrast, classification-based methods operate at the image-tile level and offer greater scalability and robustness for large, diverse, and crowdsourced UAV imagery, although at the cost of reduced spatial precision in delineating individual waste objects (Kako et al., 2026). Open data repositories such as OpenAerialMap now host large volumes of openly accessible UAV imagery, reflecting the growing promise of community-driven platforms for sharing high-resolution geospatial data, with increasing potential to broaden representation across environmental, cultural, and infrastructural contexts as contributions continue to grow. Yet existing studies predominantly emphasize methodological advances or site-specific case studies, with limited attention to cross-site generalizability, operational deployment or openly accessible models that can be readily applied to heterogeneous UAV datasets (Abdu & Mohd, 2022).

The overarching aim of this study is to provide an openly accessible, deployment-ready deep learning model for detecting ODD-MSW in crowdsourced UAV imagery from Sub-Saharan Africa. To make the scope of this contribution explicit, we pursue five specific objectives:

- **O1** To fine-tune a pre-trained, open-source image-classification model to distinguish ODD-MSW from background in high-resolution UAV image tiles (Section 2.3).
- **O2** To quantify cross-regional detection performance across 29 study areas in 10 Sub-Saharan African countries spanning diverse environmental, settlement and agro-ecological contexts (cf. Fig. A.1).
- **O3** To test transferability explicitly, by applying the frozen model without any site-specific retraining or threshold re-calibration to independent UAV scenes located outside the 29 training regions and largely outside Sub-Saharan Africa (Section 2.5).
- **O4** To establish the imagery requirements of the task by benchmarking the UAV-based approach against openly available satellite imagery, using an identical model architecture, hyperparameter configuration and evaluation protocol (Section 2.7).

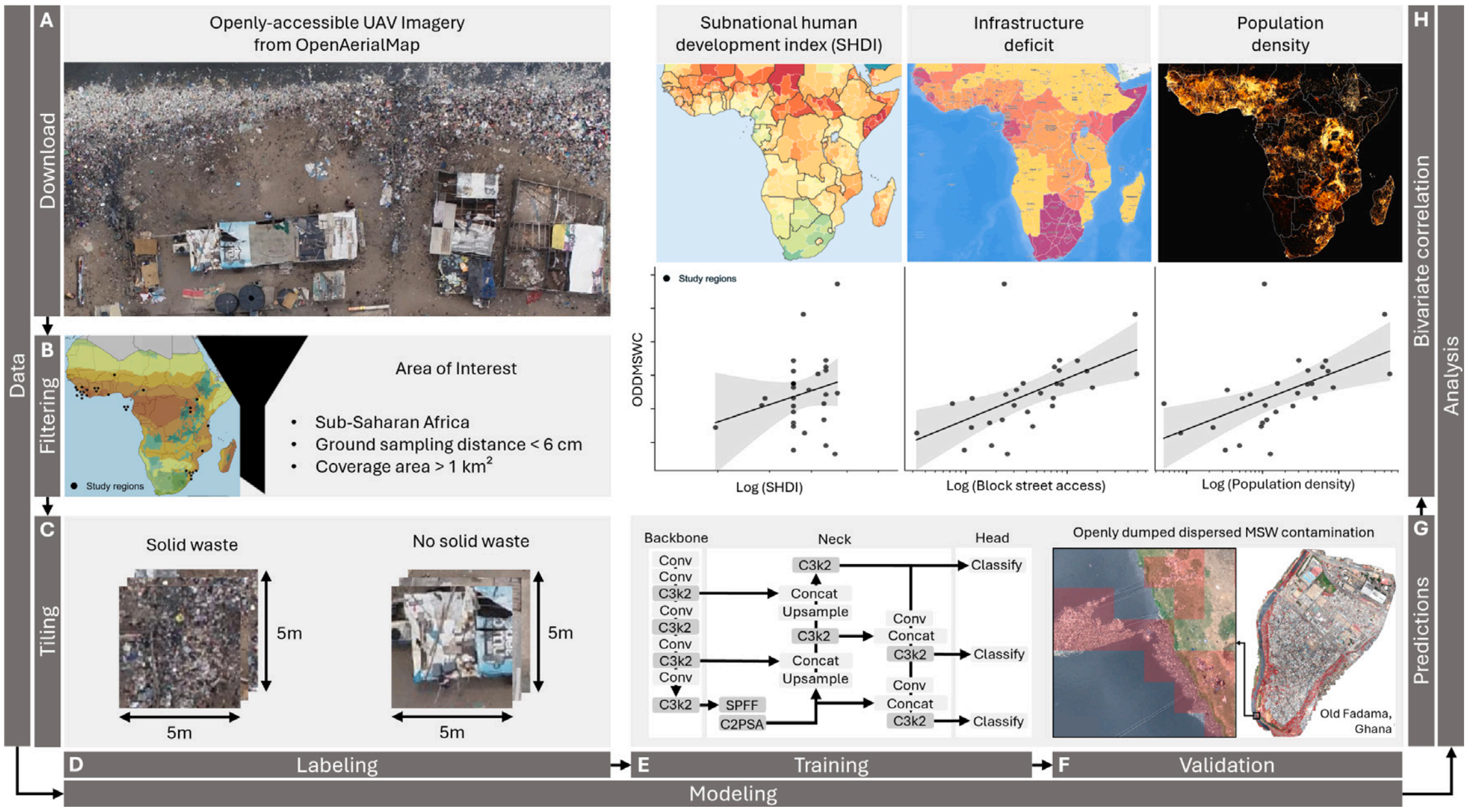

**Fig. 1.** Overview of the end-to-end workflow for large-scale UAV-based detection of ODD-MSW and socio-spatial assessment across Sub-Saharan Africa. The pipeline integrates multi-source geospatial data acquisition (A–B), pre-processing (C–D), supervised training (E), regional prediction and validation (F–G), and bivariate correlation analysis (H).

**O5** To demonstrate the operational utility of the resulting predictions by mapping fine-scale ODD-MSW distribution and quantifying its association with socio-spatial indicators of development, service accessibility and residential density (Section 2.8).

Objectives O3 and O4 distinguish this work from site-specific case studies: rather than inferring generalization from the heterogeneity of the training data alone, we measure it directly on held-out geographies and against an alternative imagery source. By prioritizing cross-site generalization and deployment readiness over methodological novelty, the framework is designed to empower local communities, authorities, researchers, and civic actors to expand UAV data collection and independently integrate newly acquired imagery into the analysis workflow, enabling iterative MSW detection and the progressive refinement of local monitoring efforts. The resulting high-resolution spatial information can support more targeted interventions and contribute to mitigating associated environmental and public health risks, including drainage obstruction and flood amplification (Mokuolu et al., 2022; Moritz & Knoblauch, 2023).

## 2. Materials and methods

To detect and quantify ODD-MSW across heterogeneous urban and peri-urban environments in Sub-Saharan Africa, we developed a multi-stage workflow encompassing high-resolution UAV imagery acquisition, pre-processing, manual labeling, and deep learning-based model training and evaluation. The overall pipeline is designed to handle large-scale UAV imagery while preserving fine spatial detail, enabling cross-regional analysis (cf. Fig. 1).

### 2.1. Study regions and UAV imagery

UAV imagery was curated from OpenAerialMap covering multiple Sub-Saharan countries. We selected datasets that met strict quality criteria, including a ground sampling distance below 6 cm—typically achievable with commercially available UAV platforms at survey flight altitudes of approximately 80–120 m—and a minimum spatial coverage of 1 $km^2$ per region. The resulting dataset comprises 29 regions across 10 countries, capturing diverse urban and peri-urban environments (cf. Fig. A.1, Table A.1). Liberia (nine regions) and Mozambique (six) contribute the largest shares, followed by São Tomé and Príncipe and Côte d'Ivoire (three regions each). Tanzania and Ghana contribute two regions, whereas Uganda, Niger, Kenya and Sierra Leone each contribute one region. Across all selected regions, the average ground sampling distance (GSD) is 4.71 cm per pixel (range: 3.52–5.92 cm), while the mean coverage area per region is 18.26 $km^2$ (range: 1.30–77.47 $km^2$). All orthomosaics are three-band 8-bit RGB rasters; no additional spectral bands (e.g. near-infrared) are distributed with the OpenAerialMap datasets used here, and GSD is reported as the native pixel ground dimension of each orthomosaic rather than as a nominal display resolution. Complete scene-level metadata — including approximate scene-centroid coordinates, ISO 8601 acquisition dates, OpenAerialMap identifiers and data providers — are reported in Table A.1. This multi-country, high-resolution dataset enables cross-context evaluation and supports the assessment of model generalizability across diverse socio-environmental, infrastructural, and ODD-MSW accumulation conditions.

### 2.2. Tiling and manual annotation

Each imagery region was processed by generating a regular grid of 5 m × 5 m tiles using Python and the packages geopandas and rasterio, providing a uniform spatial framework for downstream analysis (cf. Fig. A.2). Tiles were extracted directly from the source GeoTIFFs using rasterio's windowed reading functionality, preserving full spatial fidelity. Tile size and sampling numbers were determined empirically through preliminary experiments to balance annotation effort and model performance. Manual visual annotation was then performed for each region, with approximately 100 ODD-MSW tiles and 100 background tiles per region, resulting in a balanced dataset of 5800 labeled tiles. ODD-MSW tiles included visible plastic litter, debris, rubber tires, informal dumping areas, and litter concentrations, whereas background tiles consisted of buildings, roads, vegetation, water, bare soil, and stones, including features that could visually resemble ODD-MSW (cf. Fig. A.3). Dataset splits were defined as 70% for training, 15% for validation, and 15% for testing to ensure unbiased model evaluation.

### 2.3. Model architecture and fine-tuning

For model development, we finetuned the YOLO11x-cls architecture from Ultralytics, comprising 28.3 million parameters across 176 layers and pre-trained on ImageNet. Input tiles were resized to 128 × 128 pixels, retained in their native three-band 8-bit RGB encoding, and scaled per channel from the integer range $[0, 255]$ to the unit interval $[0, 1]$ following the default Ultralytics pre-processing. The network was finetuned for 150 epochs with early stopping using the AdamW optimizer and an initial learning rate of 0.0005. Batch size was set to 64, and training was conducted on a single NVIDIA A100 GPU. To improve model robustness, extensive data augmentation was applied, including random horizontal and vertical flips, rotations of ±15°, brightness and contrast adjustments, and Mosaic augmentation.

### 2.4. Within-domain model evaluation

Model performance was evaluated on a held-out test dataset consisting of unseen UAV image tiles sampled from spatially and environmentally heterogeneous contexts across all study regions, enabling evaluation under diverse real-world conditions and assessment of generalization across varying settlement and environmental settings. This heterogeneous sampling strategy was preferred over leave-one-region-out or leave-one-country-out validation, as administrative boundaries do not reliably capture the relevant visual and environmental variability for this task, and substantial heterogeneity exists within individual regions. Performance was assessed using precision, recall, F1-score, receiver operating characteristic area under the curve (ROC AUC), and overall accuracy (cf. Eqs. (1)–(4)). Precision quantifies the proportion of predicted ODD-MSW tiles that were correctly classified, whereas recall measures the proportion of actual ODD-MSW tiles that were successfully detected. The F1-score, defined as the harmonic mean of precision and recall, was used as the primary evaluation metric because it balances omission and commission errors in ODD-MSW detection. ROC AUC was used to assess the model's discrimination capability across classification thresholds, while overall accuracy quantified the proportion of correctly classified tiles across both classes.

$$\text{Precision} = \frac{TP}{TP + FP} \tag{1}$$

$$\text{Recall} = \frac{TP}{TP + FN} \tag{2}$$

$$\text{F1} = 2 \times \frac{\text{Precision} \times \text{Recall}}{\text{Precision} + \text{Recall}} \tag{3}$$

$$\text{Accuracy} = \frac{TP + TN}{TP + TN + FP + FN} \tag{4}$$

where (TP), (TN), (FP), and (FN) denote true positives, true negatives, false positives, and false negatives, respectively.

### 2.5. Independent transferability assessment outside the training regions

The evaluation described in Section 2.4 quantifies performance on unseen tiles drawn from the same 29 regions used for training, and therefore measures within-domain generalization rather than transferability in the strict sense. Because sampling imagery from several countries does not by itself demonstrate that a model transfers, we additionally assembled a fully independent evaluation dataset from OpenAerialMap scenes that were not used at any stage of model development, site selection, hyperparameter tuning or threshold calibration.

Scene selection applied the same criteria as the training data (GSD below 6 cm, minimum coverage of 1 km$^2$) but deliberately targeted geographies outside the 29 training regions. The resulting dataset comprises 25 orthomosaics from 21 cities in 19 countries, distributed across Africa (2 scenes), Asia (10), Europe (3), North America (2) and South America (8). Only two scenes are located in Sub-Saharan Africa, and neither of them in a country represented in the training data, so the assessment spans climatic, morphological and waste-management settings substantially different from the training domain, including arid, temperate and continental environments absent from the original study design. For each scene, tiles were annotated by an analyst who was not involved in producing the training labels, following the labeling protocol of Section 2.2 without modification. This yielded 4794 independent reference labels (1774 ODD-MSW and 3020 background; cf. Table A.3). In contrast to the deliberately balanced training set, the positive class accounts for only 37.0% of these labels, so the transfer evaluation is also more representative of the class imbalance encountered in operational settings.

Crucially, the model weights, the input pre-processing and the 0.5 decision threshold were left unchanged: no fine-tuning, domain adaptation, per-scene threshold optimization or label re-definition was performed. The metrics reported in Section 3.3 therefore describe zero-shot transfer performance and constitute a direct test of the deployment-readiness claim (objective O3).

### 2.6. Large-scale inference and the ODD-MSWC indicator

During inference, the finetuned model was applied to all study regions using the same preprocessing. Each tile was assigned a predicted class (ODD-MSW or background) and an associated confidence score, which were stored in the corresponding spatial grid. For regional aggregation, tile-level predictions were summarized by calculating the proportion of tiles classified as ODD-MSW relative to the total number of analyzed tiles, multiplied by 100 (cf. Eq. (1)) We term this metric the Openly Dumped Dispersed MSW Contamination (ODD-MSWC), a spatially explicit metric quantifying the spatial prevalence of ODD-MSW. The ODD-MSWC enables consistent comparisons across heterogeneous environmental and urban contexts, facilitating quantitative cross-regional assessments of ODD-MSW accumulation patterns.

$$\text{ODD-MSWC}_r = \frac{N_r^{\text{MSW}}}{N_r^{\text{total}}} \times 100 \tag{5}$$

where:

$r$ is the region index

$N_r^{\text{MSW}}$ is the number of tiles classified as ODD-MSW in region $r$

$N_r^{\text{total}}$ is the total number of tiles in region $r$

ODD-MSWC is computed exclusively for the 29 Sub-Saharan African study regions listed in Table A.1, that is, for the regions used to train and evaluate the model. It is deliberately not computed for the independent scenes of Section 2.5, which serve solely to quantify detection accuracy outside the training domain and are not intended to support comparison of waste prevalence between regions. Because ODD-MSWC is defined as a proportion rather than a count, it is by construction invariant to the differing number of tiles per region. Two caveats nevertheless qualify its cross-regional comparability. First, the denominator $N_r^{\text{total}}$ is set by the extent of the uploaded orthomosaic rather than by an administrative or functional boundary, so a region whose imagery happens to include a large share of undeveloped land will yield a lower ODD-MSWC than an otherwise comparable region imaged only over its built-up core; ODD-MSWC is thus a property of the imaged scene, not of the settlement as a whole. Second, the visual appearance of ODD-MSW differs between regions as a function of waste composition, local disposal practice, substrate and illumination, which could in principle induce region-specific detection bias and hence bias the indicator. We addressed this by drawing training tiles from all 29 regions rather than from a single site, by applying extensive photometric and geometric augmentation (Section 2.3), and by reporting per-region evaluation metrics (Table A.2) so that the contribution of any individual region to aggregate performance can be verified directly. The narrow spread of per-region F1 scores indicates that no single region dominates the aggregate result. Residual limitations of the indicator, including its dependence on tile size and decision threshold, are discussed in Section 4.

### 2.7. Benchmark against openly available satellite imagery

To establish whether openly available satellite imagery could substitute for UAV imagery in this task (objective O4), we repeated the detection experiment over the Msimbazi delta in Dar es Salaam, Tanzania, using Bing Maps aerial imagery acquired on 16 October 2025 at the highest available zoom level (22). The results of this experiment are reported in Section 3.4.

Spatial resolution and GSD are distinct concepts and are therefore reported separately here. At zoom level 22 the Web Mercator tile pyramid has a nominal pixel spacing of 3.73 cm at the equator, corresponding to 3.71 cm at the latitude of the study area (−6.81°). This figure describes the display grid only. The native GSD of the source imagery composited into the Bing basemap over this area is approximately 50 cm, so the tiles served at zoom level 22 are upsampled from source data roughly 13 times coarser than their nominal pixel spacing, and roughly 10 times coarser than the UAV orthomosaic of the same area (5.04 cm, Table A.1). The effective information content of the satellite tiles is therefore governed by the source GSD, not by the zoom level.

To make the two experiments directly comparable, we fine-tuned the identical YOLO11x-cls architecture on the satellite tiles using exactly the configuration of Section 2.3: the same 5 m × 5 m tile geometry, 128 × 128 pixel input size, AdamW optimizer, initial learning rate of 0.0005, batch size of 64, budget of 150 epochs with early stopping, augmentation pipeline, and 70/15/15 train/validation/test split. Labels were transferred from the co-located UAV annotations, so the reference data are identical. Both datasets are three-band 8-bit RGB rasters and no further spectral bands are available for either source, so band configuration is identical across the two experiments and cannot account for any performance difference. No class-rebalancing procedure was applied to either dataset, so that the two pipelines remain symmetric in this respect as well. The imagery source and its effective GSD therefore remain the only systematic difference between the two experiments. For completeness we additionally report the performance of the unmodified UAV-trained model applied directly to the satellite tiles, which isolates the domain gap from the effect of retraining.

### 2.8. Socio-spatial covariates and statistical analysis

To explore potential socio-spatial drivers of observed ODD-MSW patterns, we analyzed bivariate correlations between the ODD-MSWC and three contextual indicators: (i) the Subnational Human Development Index (SHDI) for 2022 as a proxy for socio-economic development, defined as a composite subnational index combining normalized dimensions of life expectancy, education, and income, with values ranging from 0 to 1 (Smits & Permanyer, 2019); (ii) an infrastructure deficit index capturing the lack of road access to buildings relevant to MSW management services (Bettencourt & Marchio, 2025), defined as the proportion of buildings in each spatial unit that are not within a defined accessibility threshold distance to the nearest mapped road segment, where higher values indicate lower accessibility; and (iii) gridded population density estimates from the 2025 WorldPop dataset as an indicator of potential household waste generation, derived from 100 m resolution population surfaces and aggregated to persons per square kilometer at the study-region level (WorldPop, 2026). Bivariate correlation analyses were conducted at the study-region level using spatially aggregated indicator values matched to each region's spatial extent. This approach enabled us to assess whether variation in ODD-MSW prevalence is systematically associated with differences in development status, service accessibility and residential density across heterogeneous contexts.

The covariates and the imagery are not contemporaneous. SHDI is used for the reference year 2022 and WorldPop for 2025, whereas the UAV imagery spans 2016–2025 (Table A.1), so for the oldest scenes the covariates post-date the imagery by up to nine years. Year-matched values could not be substituted: subnational SHDI estimates are not published for every region-year in the sample, and successive WorldPop releases differ methodologically to an extent that makes a mixed-vintage population surface internally inconsistent. We therefore treat both covariates as time-invariant region descriptors, interpret all reported associations as cross-sectional rather than causal, and report a sensitivity analysis restricted to the 19 scenes acquired from 2020 onwards, for which the temporal offset is at most five years. This constraint is stated explicitly as a limitation in Section 4.

Associations were quantified using Spearman's rank correlation coefficient $\rho$, which does not assume linearity or normality and is robust to the skewed indicator distributions and small sample ($n = 29$ regions). For every pairwise association we report $\rho$, the two-sided $p$-value, a 95% confidence interval obtained by bias-corrected and accelerated bootstrap over regions (10,000 resamples), and $\rho^2$ as a rank-based measure of shared variance, so that statistical significance and effect size can be assessed separately. Statistical significance was evaluated at $\alpha = 0.05$; because five correlations are reported, we additionally give Holm–Bonferroni adjusted $p$-values, and interpretation is based on the adjusted values. Given the small number of regions, confidence intervals rather than point estimates are used as the primary basis for interpretation, and no inferential claims are made about individual regions.

## 3. Results

### 3.1. Within-domain detection performance

The finetuned model demonstrated high accuracy in classifying ODD-MSW and background tiles across the test dataset ($n = 870$). Overall, the model achieved an accuracy of 92.87%, with per-class F1 scores of 92.69% for ODD-MSW and 92.99% for background (see Fig. 2). Across individual study regions, ODD-MSW F1 scores ranged from 0.783 to 1.000 (SD: 0.054; cf. Table A.2), indicating consistently strong performance with only minor variation between heterogeneous contexts. Correct predictions were made with a mean confidence of 93.25%, while incorrect predictions exhibited lower confidence (mean 71.92%). These results indicate that the model is highly effective in distinguishing visible ODD-MSW from surrounding urban and peri-urban backgrounds under diverse environmental conditions.

### 3.2. Visual inspection of predictions

To further evaluate model performance, we conducted a visual inspection of predicted classifications across representative tiles (cf. Fig. 3). True positive predictions predominantly captured exposed ODD-MSW accumulations, such as informal dumping areas and littered streets. True negative tiles consisted of typical background features, including buildings, roads, vegetation, bare soil, and objects resembling ODD-MSW, for example, drying clothes or patterned surfaces. Common sources of false positives included sunlight reflections on water surfaces, as well as small piles of rubble or debris resembling ODD-MSW. Notably, while rubble and debris may be classified as ODD-MSW in certain contexts — particularly when discarded away from active construction or demolition sites — they were excluded from our labeling criteria to maintain taxonomic consistency. Consequently, these materials are categorized as false positives within this specific study. False negatives were generally observed where ODD-MSW was hidden, such as beneath tree canopies, shelters, or within dense piles of discarded tires. Although discarded tires may constitute MSW in other settings, they were omitted from our target class due to their frequent repurposing in the Sub-Saharan context; their omission here thus represents a consistent application of our localized labeling schema. Collectively, these instances elucidate the environmental and anthropogenic factors that define the model's operational boundaries, providing essential qualitative depth to the quantitative performance metrics.

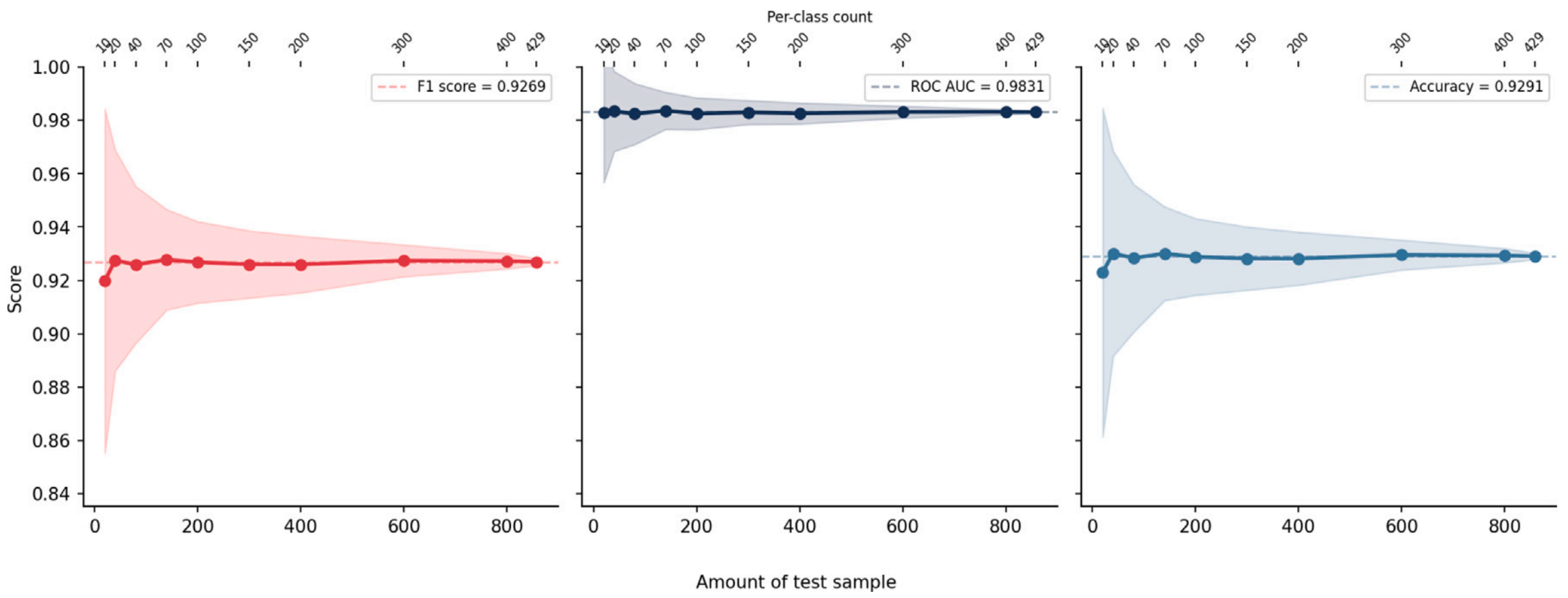

**Fig. 2.** Stability of evaluation metrics under stratified bootstrap subsampling of the test set. Performance metrics were evaluated across increasing test-set sizes using stratified bootstrap subsampling that preserved the class balance between ODD-MSW and background tiles. Curves show the mean value of each metric over bootstrap replicates, with shaded regions indicating variability across resamples. (Left) F1 score as a function of total test samples. (Center) Receiver operating characteristic area under the curve (ROC AUC). (Right) Classification accuracy. Points along the $x$-axis correspond to increasing numbers of total test samples (ODD-MSW + background), with the upper axis indicating the corresponding per-class sample count. Across all metrics, performance rapidly stabilizes as sample size increases, converging near F1 $\approx$ 0.927, ROC AUC $\approx$ 0.983, and accuracy $\approx$0.929, indicating that the reported model performance is robust to moderate variations in test-set composition.

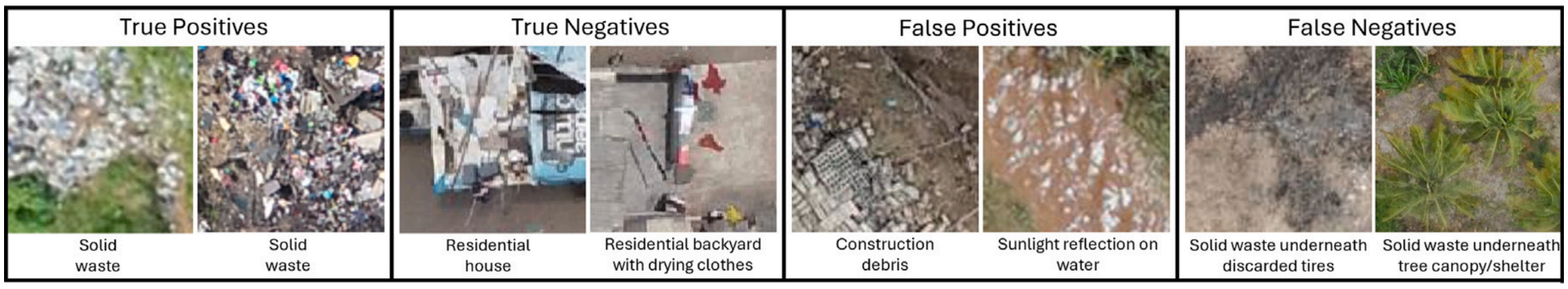

**Fig. 3.** Visual inspection of model predictions. Representative examples of true positives, true negatives, false positives, and false negatives are shown.

### *3.3. Transferability to independent scenes outside the training regions*

Applied without any fine-tuning, threshold re-calibration or other adaptation to the 25 independently annotated out-of-sample scenes (Section 2.5), the model classified the 4794 reference tiles with an overall accuracy of 89.4%, an ODD-MSW precision of 0.861, a recall of 0.874, an F1 score of 0.867 and a ROC AUC of 0.951. Relative to the within-domain test set (F1 = 0.927), zero-shot transfer therefore costs approximately 6 F1 points—a moderate and, for an operational screening tool, acceptable degradation given that the evaluation set is drawn from 19 countries on five continents, is more strongly class-imbalanced (37.0% positive) than the training data, and includes climatic and morphological settings entirely absent from training.

Performance was consistent rather than uniform across scenes. Per-scene F1 ranged from 0.718 to 0.968 with a median of 0.861 and a standard deviation of 0.061 across the 25 scenes (cf. Table A.3). The strongest results were obtained in dense, warm-climate settings resembling the training domain, including Les Cayes (Haiti; F1 = 0.968) and Kanifing (The Gambia; F1 = 0.941), the latter confirming that transfer within Sub-Saharan Africa but to an unseen country is essentially loss-free. The weakest results occurred in Lviv (Ukraine; F1 = 0.718) and Izmir (Turkey; F1 = 0.742), both temperate European scenes with low positive-class prevalence, where a small absolute number of false positives on paved and construction surfaces depresses precision disproportionately. Scenes with fewer than 100 reference labels show the widest spread, consistent with sampling variability rather than a systematic failure mode.

Taken together, these results substantiate the claim that the released model can be applied directly to newly uploaded UAV imagery without site-specific retraining, and extend that claim beyond the Sub-Saharan African contexts represented in the training data. They also delimit it: the largest performance losses arise in environments whose built form and waste morphology differ most from the training distribution, and users working in such settings should expect precision to be the limiting quantity.

### 3.4. Benchmark against openly available satellite imagery

Over the Msimbazi delta in Dar es Salaam, the UAV-based model achieved an F1 score of 0.933 (Table A.2). Applying that same model unchanged to the co-located Bing satellite tiles reduced the F1 score to 0.144 (precision 0.108, recall 0.214, accuracy 0.612, ROC AUC 0.571), confirming that the two imagery sources are not interchangeable at inference time.

Fine-tuning the identical YOLO11x-cls architecture directly on the satellite tiles, using the same hyperparameters, augmentation pipeline, tile geometry and reference labels as the UAV model (Section 2.7), improved matters only marginally: overall accuracy reached 0.673 with an ODD-MSW precision of 0.229, a recall of 0.196, an F1 score of 0.211 and a ROC AUC of 0.608. Because architecture, hyperparameters, band configuration and reference labels are held constant, the residual gap of approximately 0.72 F1 points relative to the UAV model is attributable to the effective GSD of the imagery rather than to the choice of classifier.

The failure mode is informative. Recall remains low because dispersed litter, typically a few tens of centimeters across, is smaller than a single source pixel at a native GSD of 50 cm and therefore contributes only a fraction of the mixed spectral signal of the pixel it occupies; upsampling the tiles to a nominal 3.71 cm pixel spacing interpolates this signal but adds no information. Precision remains low because, at this resolution, litter, gravel, damaged roofing and bare soil become spectrally and texturally indistinguishable.

We applied the trained model across all study regions to generate spatially explicit predictions of ODD-MSW distribution. Predictions were generated for more than 13 million 5 m × 5 m imagery tiles, enabling high-resolution characterization of localized contamination across study regions (cf. Fig. 4). The resulting maps reveal pronounced spatial heterogeneity in ODD-MSW prevalence. While some urban centers exhibit more extensive contamination than others, clear differences also emerge in the spatial configuration of ODD-MSW distribution across study regions. In certain regions, ODD-MSW is relatively evenly dispersed across the landscape. In contrast, regions characterized by waterways or heterogeneous built environments show more spatially structured patterns, with ODD-MSW concentrated in clearly delineated hotspots and aligned along specific landscape features (cf. Fig. 5).

### 3.5. Regional ODD-MSW distribution and association with socio-spatial indicators

Reflecting these observations, ODD-MSWC values reveal substantial variation across study sites. The highest ODD-MSWC was observed in Old Fadama, a community area in Accra, Ghana, characterized by dense built-up areas and higher population density, and associated with a major market. Lower values were primarily recorded in peri-urban settings of Monrovia in Liberia and Kampala in Uganda (cf. Fig. 4). These spatial differences are consistent with the observed associations between ODD-MSWC and contextual characteristics: higher contamination levels are positively correlated with population density ($\rho = 0.674$, 95% CI [0.41, 0.83], $p < 0.001$) and infrastructural constraints, as captured by the infrastructure deficit index ($\rho = 0.700$, 95% CI [0.45, 0.85], $p < 0.001$), which was introduced as a higher-resolution alternative to SHDI (Fig. 6). In contrast, no clear association is observed with SHDI in the bivariate analysis ($\rho = 0.215$, 95% CI [−0.16, 0.54], $p = 0.263$).

Full inferential statistics for all five reported correlations are given in Table 1. The associations with the infrastructure deficit index and with population density remain statistically significant after Holm–Bonferroni adjustment for the five comparisons (adjusted $p < 0.001$ in both cases) and correspond to large effect sizes, with 49.0% and 45.4% of rank variance shared, respectively. The association with SHDI is not significant and its confidence interval spans zero, so the data are compatible with anything from a weak negative to a moderate positive association; the appropriate conclusion is that this sample cannot resolve an SHDI effect, not that none exists. Among the predictors, population density and the infrastructure deficit index are moderately correlated ($\rho = 0.632$, $p < 0.001$) while the infrastructure deficit index and SHDI are effectively unrelated ($\rho = 0.086$, $p = 0.657$), indicating no substantial collinearity. The sensitivity analysis restricted to the 19 scenes acquired from 2020 onwards leaves the pattern unchanged (infrastructure deficit $\rho = 0.681$, population density $\rho = 0.643$, SHDI $\rho = 0.188$), indicating that the temporal offset between imagery and covariates (Section 2.8) does not drive the reported associations. Given $n = 29$ regions, all confidence intervals are wide and the estimates should be read as indicative of the direction and approximate magnitude of association rather than as precise effect estimates.

Together with other factors not explicitly captured in this study, such as local MSW management practices and infrastructure provision, these variables may contribute to the observed magnitude and spatial configuration of ODD-MSW accumulation.

## 4. Discussion

High-resolution UAV imagery combined with a fine-tuned deep learning model enables the detection of ODD-MSW across diverse urban and peri-urban environments in Sub-Saharan Africa. The consistently high F1 scores, with only minor variation across study regions, demonstrate strong cross-context generalization. This within-domain consistency alone would not establish transferability, since all 29 regions contributed training tiles. The zero-shot evaluation on 25 independently annotated scenes from 19 countries (Section 3.3) provides the missing direct evidence: the frozen model retained an F1 score of 0.867 without any site-specific retraining, threshold re-calibration or label re-definition. This is a critical prerequisite for operational deployment and supports the claim that the model can be applied to newly available OpenAerialMap data without requiring site-specific retraining. The transfer results also bound that claim: performance declined by roughly 6 F1 points overall and by up to 21 points in individual temperate, low-prevalence scenes, so practitioners working in settings that differ markedly from Sub-Saharan urban and peri-urban environments should expect reduced precision and are advised to verify a sample of predictions before acting on them.

Despite this strong performance, several limitations remain that delineate the current operational boundaries of the model. A primary challenge arises from objects that are visually similar to MSW but are not consistently classifiable as such. These include, for example, patterned or mosaic ground surfaces, drying clothes, or construction-related materials such as gravel and rubble. The classification of such objects is inherently context-dependent: materials like gravel may represent active construction in one setting but constitute discarded waste in another. Similarly, items such as used tires may function as planters or infrastructure elements (e.g., road markings) in some contexts, while representing openly dumped waste in others. The current tile-based classification approach lacks the semantic and contextual awareness required to reliably disambiguate these cases, leading to systematic false positives in visually ambiguous scenarios.

A second key limitation concerns false negatives, particularly in cases where MSW is partially or fully occluded. Common examples include waste concealed beneath vegetation (e.g., mangroves, palms), or roofing elements. Since the model operates on isolated image tiles without incorporating broader spatial context, it is inherently constrained in its ability to infer the presence of hidden waste.

Future work could address these limitations by incorporating spatially context-aware modeling approaches. Rather than treating each tile independently, predictions could be conditioned on the surrounding spatial neighborhood, enabling the model to capture spatial autocorrelation patterns characteristic of waste accumulation. For instance, the presence of ODD-MSW in adjacent tiles could increase the likelihood of waste occurrence in partially occluded areas. This would allow predictions to be expressed probabilistically, reflecting not only direct visual evidence but also contextual cues from the local environment. Additionally, integrating semantic scene understanding — such as the detection of vegetation types, built structures, or land-use patterns — could further improve the model's ability to distinguish between ambiguous object classes and context-dependent waste representations. Importantly, such contextual information may also encode structural constraints on dumping behavior itself: densely vegetated areas, particularly those with limited accessibility, are likely to exhibit lower probabilities of open dumping, whereas open and accessible environments — such as riverbeds, drainage corridors, or roadside areas — may systematically facilitate waste accumulation. Incorporating these environmental affordances into the modeling framework could therefore improve both detection accuracy and the interpretability of predicted spatial patterns.

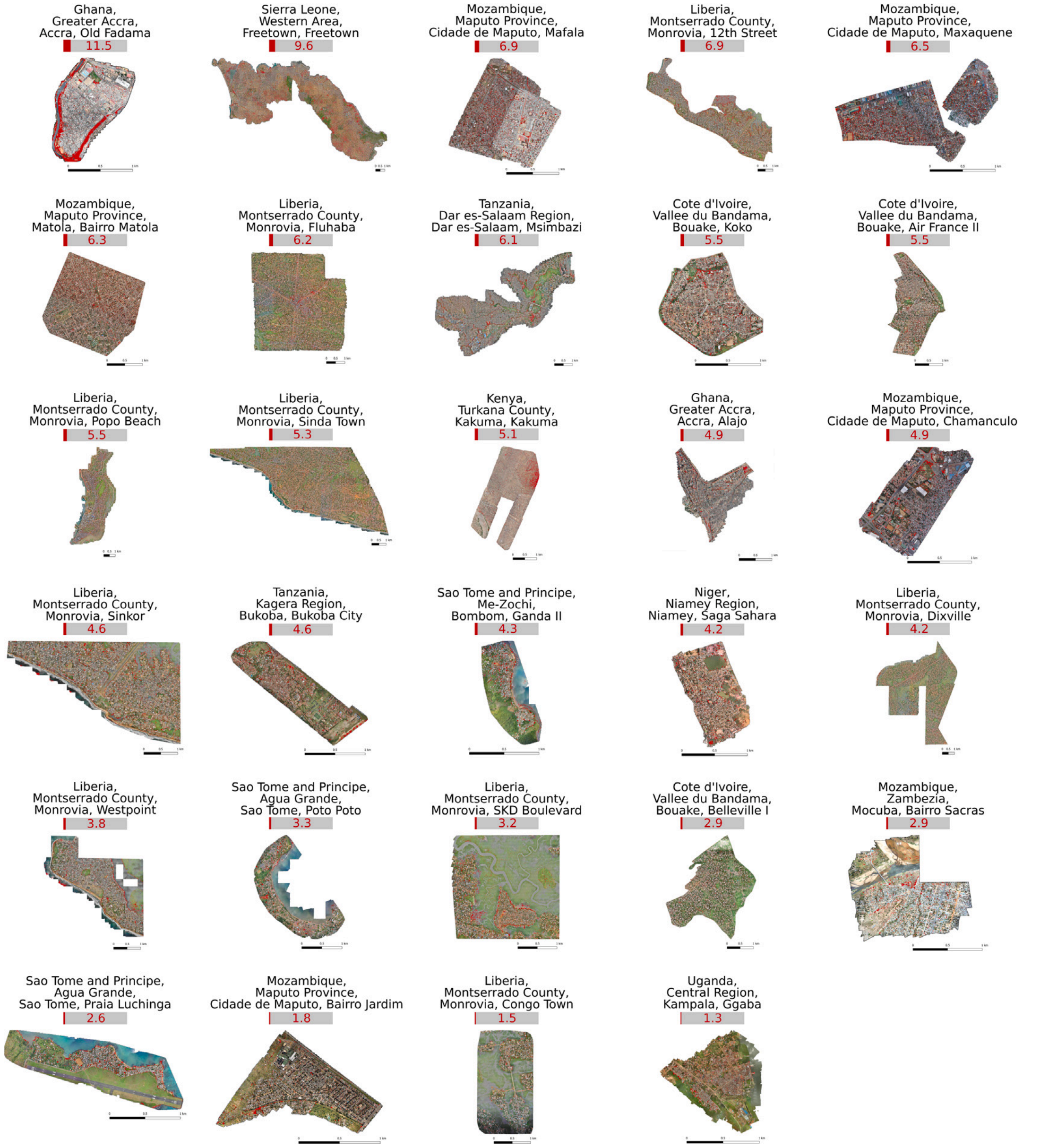

**Fig. 4.** Predicted ODD-MSW distribution across study sites. Each panel represents one of 29 regions. Red 5 m × 5 m grid cells indicate model-predicted ODD-MSW locations, highlighting fine-scale spatial patterns including hotspots of dumping. Accompanying bar charts show calculated ODD-MSWC values, which were used to rank the study sites. These maps provide (i) a comprehensive overview of model predictions across diverse urban and peri-urban African contexts, (ii) can support quantitative cross-regional comparisons, and (iii) provide a foundation for potential enhancement of local MSW management principles.

**Table 1**
Bivariate rank correlations between ODD-MSWC and socio-spatial indicators, and among predictors ($n$ = 29 regions). Reported are Spearman's $\rho$, the 95% confidence interval from a bias-corrected and accelerated bootstrap over regions (10,000 resamples), the two-sided $p$-value, the Holm–Bonferroni adjusted $p$-value across the five comparisons, and $\rho^2$ as a measure of shared rank variance. SHDI: Subnational Human Development Index.

| Variable 1 | Variable 2 | $\rho$ | 95% CI | $p$ | $p_{\text{Holm}}$ | $\rho^2$ |
|---|---|---|---|---|---|---|
| ODD-MSWC | Infrastructure deficit index | 0.700 | [0.45, 0.85] | $2.4 \times 10^{-5}$ | $1.2 \times 10^{-4}$ | 0.490 |
| ODD-MSWC | Population density | 0.674 | [0.41, 0.83] | $6.1 \times 10^{-5}$ | $2.4 \times 10^{-4}$ | 0.454 |
| ODD-MSWC | SHDI | 0.215 | [−0.16, 0.54] | 0.263 | 0.526 | 0.046 |
| Population density | Infrastructure deficit index | 0.632 | [0.35, 0.81] | $2.4 \times 10^{-4}$ | $7.1 \times 10^{-4}$ | 0.399 |
| Infrastructure deficit index | SHDI | 0.086 | [−0.29, 0.44] | 0.657 | 0.657 | 0.007 |

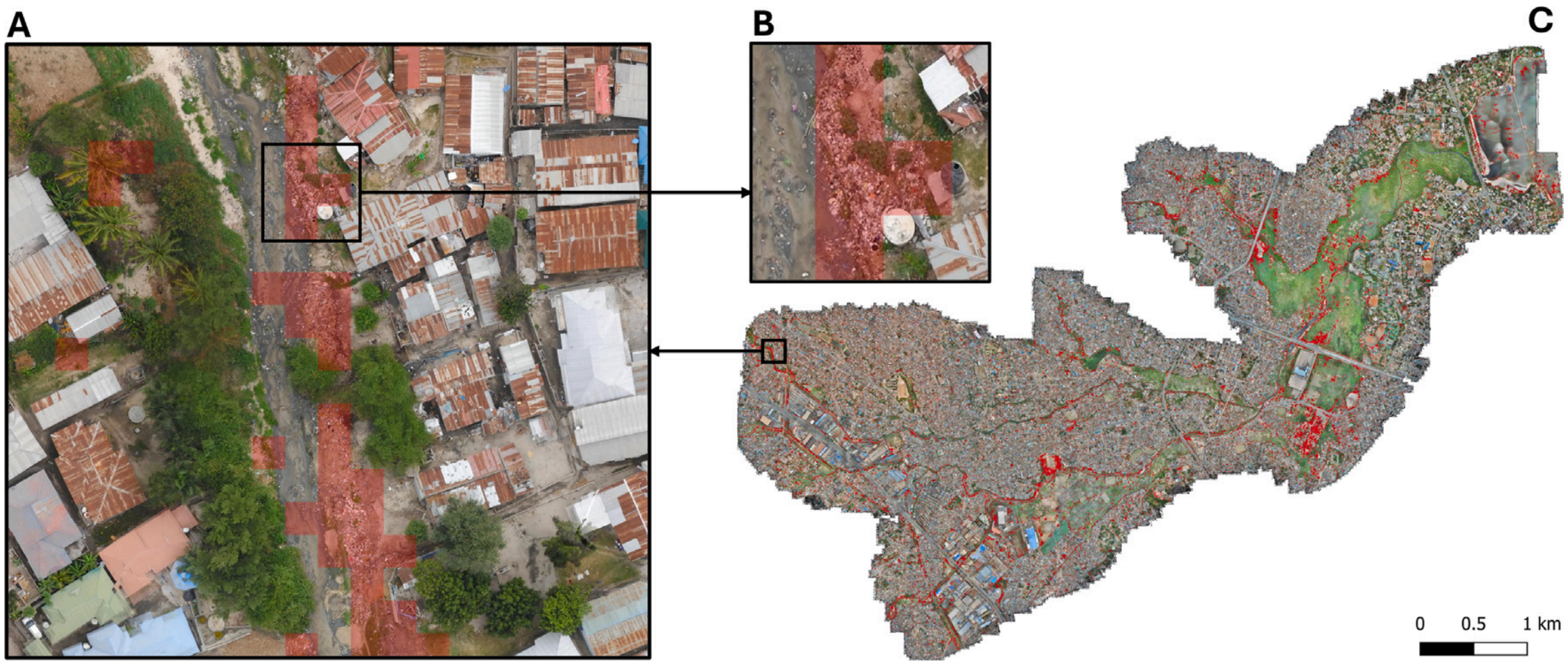

**Fig. 5.** ODD-MSW contamination in Dar es Salaam, Tanzania detected from UAV imagery. Red tiles indicate detections of ODD-MSW at 5 m × 5 m resolution. Panels A and B show zoomed-in views; panel C highlights concentrations along pathways and waterways.

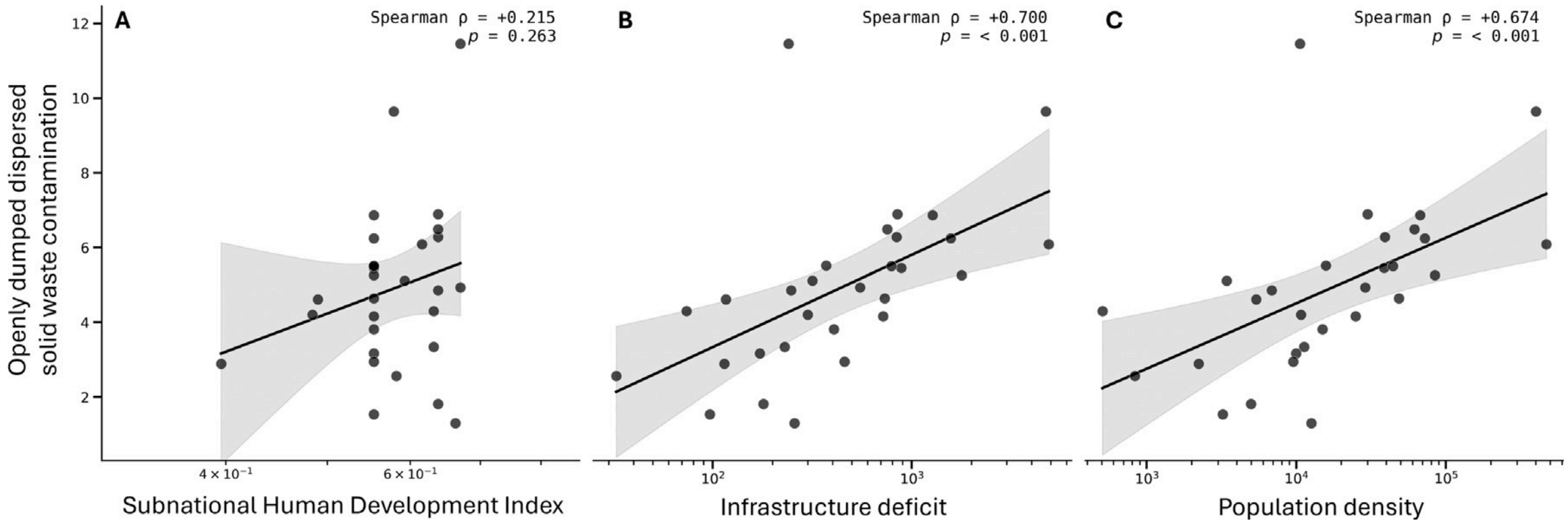

**Fig. 6.** Bivariate associations between relative solid waste contamination and socio-spatial indicators. Scatterplots show study-region–level relationships between the ODD-MSWC and three contextual indicators: SHDI (A), an infrastructure deficit index capturing the lack of road access to buildings (B), and gridded population density (C). Spearman's rank correlations indicate no significant association with SHDI ($\rho = 0.215$, $p = 0.263$), but stronger positive relationships with infrastructure deficit ($\rho = 0.700$, $p < 0.001$) and population density ($\rho = 0.674$, $p < 0.001$). Full statistics, including confidence intervals and effect sizes, are reported in Table 1. The positive association with infrastructure deficit index suggests that greater infrastructural deficits are linked to higher ODD-MSWC, whereas the positive association with population density indicates increased ODD-MSWC in more densely populated regions. Correlations among predictors were moderate between population density and infrastructure deficit ($\rho = 0.632$, $p < 0.001$) and negligible between infrastructure deficit and SHDI ($\rho = 0.086$, $p = 0.657$), indicating no substantial collinearity. The weak association between SHDI and infrastructure deficit underscores the limited explanatory relevance of SHDI at this spatial resolution and motivates the use of the infrastructure deficit metric in this context.

The choice of a 5 m × 5 m tile size reflects a compromise between capturing fine-scale heterogeneity in dispersed MSW and ensuring computational efficiency across large multi-regional UAV datasets. This discretization directly influences the ODD-MSWC metric through spatial aggregation, as alternative window sizes may alter the estimated proportion of detected MSW due to scale-dependent mixing effects. A systematic sensitivity analysis across multiple spatial resolutions therefore represents an important direction for future work to quantify the robustness of observed spatial patterns under varying aggregation schemes. In this context, ODD-MSWC should be interpreted as a tile-based measure of classified spatial coverage of ODD-MSW, rather than a direct proxy for waste abundance, density, or probabilistic contamination. As a derived indicator, it is inherently conditioned on methodological choices, including tile size, classification threshold, spatial aggregation strategy, and regional delineation. These factors jointly influence absolute values and limit direct quantitative comparability across heterogeneous study areas. Accordingly, ODD-MSWC is best understood as a relative, data-driven descriptor of spatially classified MSW occurrence within a consistent analytical framework, rather than a transferable estimate of physical waste quantity or intensity across cities or countries.

Uncertainty in model predictions is partially captured through tile-level confidence scores produced during inference, enabling the identification of high- and low-certainty predictions across heterogeneous environments. In addition, variability in performance across study regions and bootstrap resampling provides an empirical indication of prediction stability. While explicit spatial uncertainty products (e.g., entropy or probability surfaces) are not visualized in this study, such outputs can be directly derived from the existing inference framework and represent a natural extension for future work.

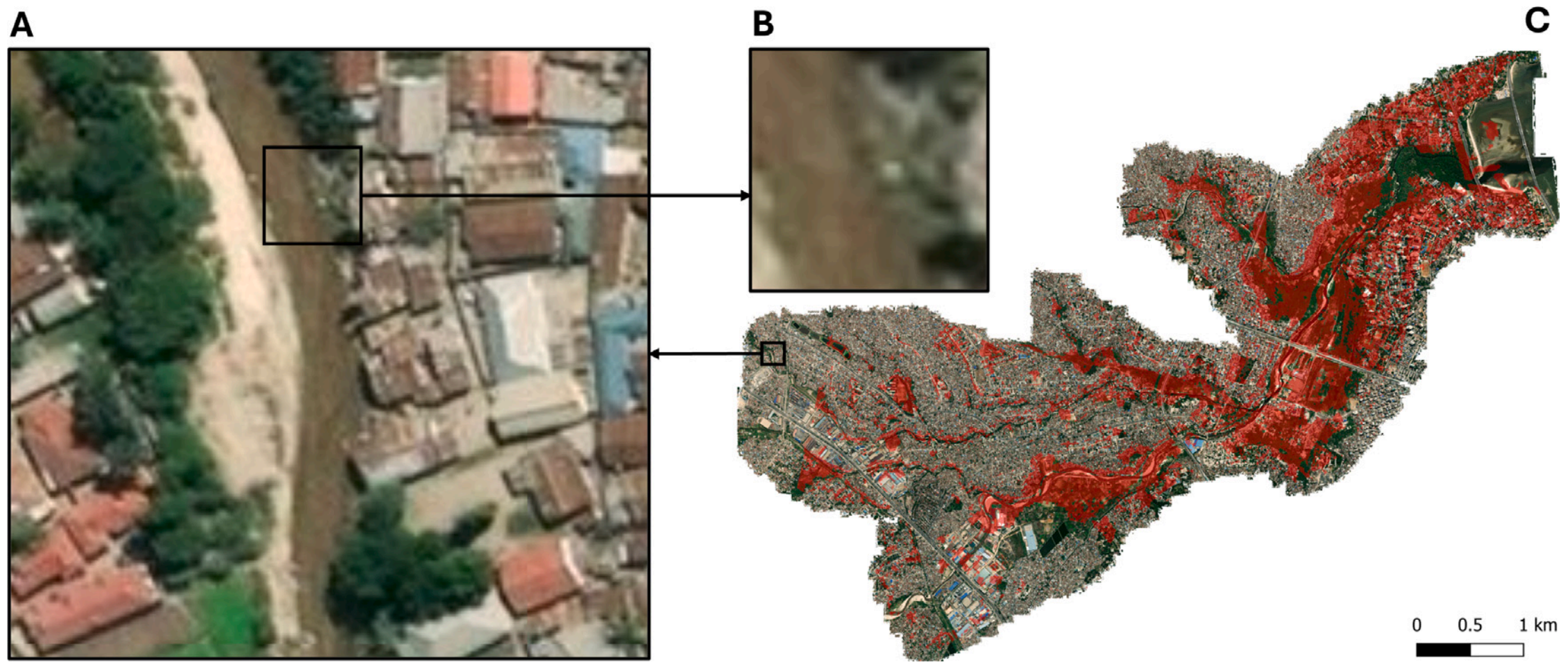

**Fig. 7.** ODD-MSW in Dar es Salaam, Tanzania, detected from openly available satellite imagery. Red tiles indicate detections of ODD-MSW. Panels A and B show zoomed-in views, while panel C provides a large-scale overview. All panels illustrate the limited spatial resolution and detection performance of openly available satellite imagery compared with UAV imagery (Fig. 5). An additional side-by-side comparison directly contrasting the spatial resolution of open-source satellite and UAV imagery is shown in Fig. A.4.

Overall, while the presented model constitutes a substantial step towards scalable, high-resolution monitoring of ODD-MSW, addressing these context sensitivity and occlusion challenges will be essential for further improving detection reliability and supporting more nuanced, decision-relevant applications in heterogeneous real-world environments.

Many of the fine-scale spatial patterns detected in this study — including dispersed litter and small informal dumping sites embedded within dense urban fabric — remain below the detection threshold of currently available open-access satellite imagery. While UAV imagery enables detection at the required spatial granularity, its applicability is constrained by limited spatial coverage, data availability, and the logistical and financial costs associated with UAV deployment. In contrast, satellite imagery offers complementary advantages, including broader spatial coverage and higher temporal revisit frequencies, thereby enabling more continuous monitoring of ODD-MSW dynamics.

The satellite benchmark reported in Sections 2.7 and 3.4 quantifies this trade-off. Because the comparison holds architecture, hyperparameters, augmentation, band configuration and reference labels constant across the two imagery sources, the remaining performance gap can be attributed to imagery characteristics rather than to modeling choices. The decisive quantity is the native GSD of the source imagery (approximately 50 cm) rather than the nominal 3.71 cm pixel spacing of the zoom-level-22 tile grid: dispersed litter is generally smaller than a single source pixel and is therefore not resolvable, irrespective of the classifier applied.

These results underscore the current limitations of openly available satellite imagery for detecting fine-scale, dispersed MSW in dense urban environments (cf. Fig. 7). Future research should therefore explore the integration of higher-quality commercial satellite data, such as PlanetScope, as well as hybrid approaches that combine UAV-derived training data with multi-resolution satellite imagery to improve scalability while retaining detection sensitivity.

Situating these results relative to earlier work is complicated by the absence of a shared class definition for openly dumped waste, which makes reported accuracies only partially commensurable. Ulloa-Torrealba et al. (2023) detected residual street waste in Medellín from very-high-resolution satellite imagery combined with GIS data, using k-means segmentation followed by a random forest classifier, and explicitly conceptualized five interpretation classes—"Sure", "Half-sure", "Not-sure", "Dispersed" and "Non-garbage"—reporting accuracies of 73.95%–95.76% for the unambiguous "Sure" class while finding the "Dispersed" and "Not-sure" categories difficult to separate. Our binary schema effectively merges their "Sure", "Half-sure" and "Dispersed" categories into a single positive class while excluding construction debris and repurposed tires, so our tile-level accuracy of 92.87% is comparable with the upper end of their single-class range rather than with their multi-class results. The comparison is nonetheless instructive in a specific way: the "Dispersed" category they found hardest to classify from satellite data is precisely the target of the present study, which independently supports our satellite benchmark (Section 3.4) and the argument for centimeter-scale UAV imagery.

Youme et al. (2021) are closer to our imagery domain, applying convolutional neural networks to UAV imagery to detect clandestine dumps in Saint-Louis, Senegal, in a directly comparable West African context. Their target class is however the discrete dump feature rather than dispersed litter, and their evaluation covers a single site, so their results speak to feasibility rather than to transferability. Our contribution is complementary in both respects: we target the dispersed fraction that lies below the threshold of feature-based detection, and we quantify performance across held-out geographies. A like-for-like comparison on a shared study area and a shared class definition would be a valuable direction for future work.

At the dataset level, AerialWaste (Torres & Fraternali, 2023) and the recent DroneWaste dataset (Morandini et al., 2026) adopt far finer material taxonomies than ours — DroneWaste annotates 20 material types across 17 solid waste dumps, each mapped to a European Waste Code — and support instance-level detection and segmentation rather than tile-level classification. Both are also drawn predominantly from European sites and from designated dump locations. The contrast clarifies the design trade-off underlying our schema: we deliberately trade taxonomic granularity for geographic breadth and for a class definition calibrated to Sub-Saharan practice, in which tires used as planters or dikes and rubble at active construction sites are not waste. A finer taxonomy would be preferable for material-recovery or enforcement applications, and mapping our binary class onto the European Waste Code hierarchy used by DroneWaste is a natural next step towards interoperable benchmarks. Conversely, the geographic breadth and zero-shot evaluation reported here address a gap that the existing datasets, however finely annotated, do not close.

Model evaluation was constrained by the size and geographic coverage of the test dataset and by human biases inherent in manual labeling. Ambiguous boundaries of ODD-MSW and diverse cultural interpretations of what constitutes MSW influenced labeling decisions (Majchrowska et al., 2022). For example, discarded tires, often used as planters, dikes, or road boundaries, were excluded from labels due to their functional utility in local contexts (Jin et al., 2023). Construction debris was similarly excluded in the labeling schema, as its classification depends strongly on context: it may represent active material on construction sites but constitutes MSW when deposited in natural environments. Efforts to mitigate such biases included consultation with local experts, but broader training datasets spanning more regions, cultural contexts, illumination conditions, scales, and ODD-MSW morphologies would further enhance generalization and reduce residual misclassifications (Chen et al., 2022). It may also be possible to classify ODD-MSW by category, such as tires, plastics, construction rubble, etc.

In addition to spatial and labeling-related constraints, the analysis does not explicitly account for temporal variability in the UAV datasets, which span multiple acquisition years across study regions. As a result, observed differences in ODD-MSW between the study sites may partially reflect temporal changes in urban development and waste management practices, rather than purely spatial variability. Future work incorporating time-stratified datasets or repeated acquisitions would be necessary to assess temporal consistency and dynamics of MSW accumulation. Relatedly, the UAV imagery used in this study spans a substantial temporal range (approximately 2016–2025; cf. Table A.1), and all regions are treated as contemporaneously comparable in the socio-spatial analysis. This temporal heterogeneity may introduce additional confounding, as underlying urban form, infrastructure conditions, and waste management systems may have evolved substantially over this period, potentially influencing cross-regional comparability of the observed associations. The socio-spatial covariates compound this issue, since they are fixed in time while the imagery is not: SHDI refers to 2022 and WorldPop to 2025, so for the earliest scenes — acquired in 2016 and 2017 — the covariates post-date the imagery by up to nine years, a period over which subnational development status and population distribution can change materially in rapidly urbanizing settings. This is a genuine limitation rather than a technicality, and it means that a covariate value may describe a socio-economic context that did not yet exist when the waste was imaged. Because year-matched covariates are not available for the full sample (Section 2.8), we could not eliminate the mismatch; we instead restricted a sensitivity analysis to the 19 scenes acquired from 2020 onwards, where the maximum offset falls to five years, and found the associations essentially unchanged (Section 3.5). The reported correlations should nonetheless be read as cross-sectional associations between an imaged scene and a broadly contemporaneous socio-spatial context, not as estimates of a temporally aligned relationship, and studies aiming at causal or longitudinal inference will require covariates matched to the acquisition year of each scene. A further limitation concerns the manually labeled dataset, which was intentionally constructed using balanced sampling of positive and negative tiles per region. This design does not reflect the naturally imbalanced distribution of ODD-MSW in real-world environments, where occurrences are comparatively rare relative to background classes. As a result, although this sampling strategy is appropriate for learning robust discriminative features across heterogeneous contexts, aggregate performance metrics such as accuracy and F1-score may not directly translate to operational settings, where model utility is more strongly governed by precision–recall trade-offs and the calibration of decision thresholds under highly skewed class prevalence.

Predicted ODD-MSW accumulation shows a positive correlation with the infrastructure deficit index, with dense built-up structure and lower road network access to buildings exhibiting higher prevalence of ODD-MSW (cf. Fig. 6). The association holds across the full range of the indicator, and the rank-based estimator used here is insensitive to the absolute magnitude of individual values. Accra, which records the highest ODD-MSWC in the sample, sits at the upper end of both distributions and is therefore consistent with the overall pattern; detection performance there is comparable to the sample as a whole (F1 = 0.933; Table A.2), indicating that its elevated ODD-MSWC reflects genuinely higher waste prevalence rather than reduced classification accuracy.

This region-level analysis is separate from the assessment of detection performance, which is evaluated at tile level and, for transferability, on independent scenes that play no part in the correlation analysis (Section 3.3). This association may be consistent with limited accessibility for formal collection services and the tendency for informal dumping in constrained urban fabrics. However, the interpretation of this observed correlation between ODD-MSWC and the infrastructure deficit index should be approached cautiously, as it may also be influenced by unobserved or confounding factors such as urban morphology, local governance structures, economic activity, differences in imagery acquisition dates, and other context-specific conditions not explicitly captured in this analysis. Within this framing, the infrastructure deficit index can be understood as a structural covariate associated with spatial variation in ODD-MSWC, which may be useful for descriptive comparison and urban planning considerations, including identifying areas where limited accessibility co-occurs with higher observed waste prevalence and where interventions such as improved connectivity, service coverage optimization, or land-use formalization could be further evaluated (Bettencourt & Marchio, 2025). By contrast, no clear trend was observed between ODD-MSW prevalence and SHDI, reflecting the coarse spatial resolution of this aggregated socio-economic measure. These findings underscore the importance of localized, high spatial resolution indicators, such as the infrastructure deficit index published by Bettencourt and Marchio (2025), which can effectively identify deficits in road access to buildings at the street-block level and guide actionable local interventions.

UAV-based ODD-MSW monitoring can strengthen both formal and informal MSW management systems, which together underpin urban sanitation in many Sub-Saharan African cities. Formal infrastructure — including large dumpsites, drainage networks, and organized collection routes — supports systematic removal of ODD-MSW, while informal solid waste pickers recover recyclables such as plastics and metals to generate income and manage localized ODD-MSW independently. Along with local knowledge, high-resolution UAV-derived ODD-MSW maps can enhance both approaches by identifying hotspots along drainage corridors, informal dumping sites, and areas underserved by formal collection due to limited access or infrastructural constraints. Importantly, much of the UAV imagery analyzed in this study originates from local communities and civic organizations, rather than government authorities. By applying the trained model to these datasets, residents can actively map ODD-MSW within their neighborhoods, identify hotspots, and prioritize areas for collection or cleanup, including holding their local service providers accountable. This approach empowers communities to coordinate and enhance their own efforts, complementing formal infrastructure while fostering local stewardship of urban sanitation. High-resolution, community-driven mapping thus transforms UAV imagery from a purely observational tool into an instrument for locally led, adaptive, and context-sensitive MSW management. Beyond strengthening local stewardship, this approach can also motivate communities to acquire UAV imagery more frequently, enabling regular, longitudinal monitoring of changes in ODD-MSW distribution and other challenges and the effectiveness of ongoing collection or cleanup efforts.

Building on the labeling approach used in this study, which considered local practices, future work could integrate community-driven field mapping to complement labels derived remotely from aerial imagery. This could improve validation by revealing ODD-MSW instances that are difficult to detect from above, potentially capturing additional false negatives. Tools such as ChatMap (Humanitarian OpenStreetMap Team, 2026), a web-based platform that converts messages from apps like WhatsApp, Telegram, or Signal into georeferenced maps, could facilitate this process. By relying on GPS signals, it can also function offline, making it accessible in regions with limited connectivity. Beyond enhancing model evaluation, such approaches may foster environmental awareness, strengthen community engagement, support advocacy, and stimulate innovations in problem solving and additional data collection by local residents. Importantly, high-resolution UAV imagery collected in this context could also benefit other urban and peri-urban applications, including monitoring infrastructure, green spaces, flood risk, or informal settlements, thereby extending the utility of these datasets well beyond MSW management.

Future work will focus on operationalizing the presented model within more user-friendly, open mapping ecosystems to maximize real-world applicability. In particular, integration into FAIR, an open AI-assisted mapping service developed by the Humanitarian OpenStreetMap Team (HOT), will enable practitioners to apply the model directly to newly acquired UAV imagery through an intuitive graphical user interface (HeiGIT, 2026). This would allow users to perform inference and, where necessary, fine-tune the model on local datasets without requiring advanced technical expertise, thereby substantially lowering barriers to adoption. Such accessibility is a critical prerequisite for achieving sustained real-world impact in resource-constrained settings. In parallel, ongoing research explores the transition from tile-based image classification towards spatially explicit pixel-level segmentation approaches. Specifically, the fine-tuning of Segment Anything Model (SAM) using multimodal prompts is expected to enable more precise delineation of ODD-MSW objects, improving both detection accuracy and interpretability. Together, these developments will enhance the scalability, usability, and analytical depth of UAV-based ODD-MSW monitoring, supporting more effective data-driven MSW management across diverse urban contexts.

## 5. Conclusion

This study demonstrates that deep learning models combined with UAV imagery can reliably reveal fine-scale ODD-MSW patterns across heterogeneous Sub-Saharan urban landscapes, capturing dispersed and informal dumping at scale often overlooked by traditional labor-intensive surveys. It also shows that a deep learning model can be trained to detect ODD-MSW robustly despite variation in vegetation, housing, or local practices. Applied without retraining to 25 independently annotated scenes from 19 countries on five continents, the model retained an F1 score of 0.867, indicating that this robustness extends beyond the contexts represented in the training data, while a controlled comparison against openly available satellite imagery under an identical architecture and training protocol shows that centimeter-scale imagery is a requirement of the task rather than a convenience. By providing an open-access model, this work seeks to empower municipalities, local mapping communities, and practitioners to transform raw UAV imagery into actionable insights without extensive technical expertise. Such capability is critical for guiding interventions, mitigating environmental and public health risks, and supporting adaptive, locally driven urban sanitation strategies where formal MSW management infrastructure is limited, effectively enhancing local capacity to monitor, plan, and respond to MSW challenges.

## CRediT authorship contribution statement

**Steffen Knoblauch:** Writing – review & editing, Writing – original draft, Visualization, Validation, Supervision, Software, Resources, Project administration, Methodology, Investigation, Funding acquisition, Formal analysis, Data curation, Conceptualization. **Ram Kumar Muthusamy:** Visualization, Validation, Software, Data curation. **Luís M.A. Bettencourt:** Writing – review & editing. **Costas Velis:** Writing – review & editing. **Pierre Chrzanowski:** Writing – review & editing. **Edward Charles Anderson:** Writing – review & editing. **Pete Masters:** Writing – review & editing. **Innocent Maholi:** Writing – review & editing. **Antonio Inguane:** Writing – review & editing. **Levi Szamek:** Writing – review & editing. **Alexander Zipf:** Funding acquisition.

## Appendix

See Figs. A.1–A.4 and Tables A.1–A.3

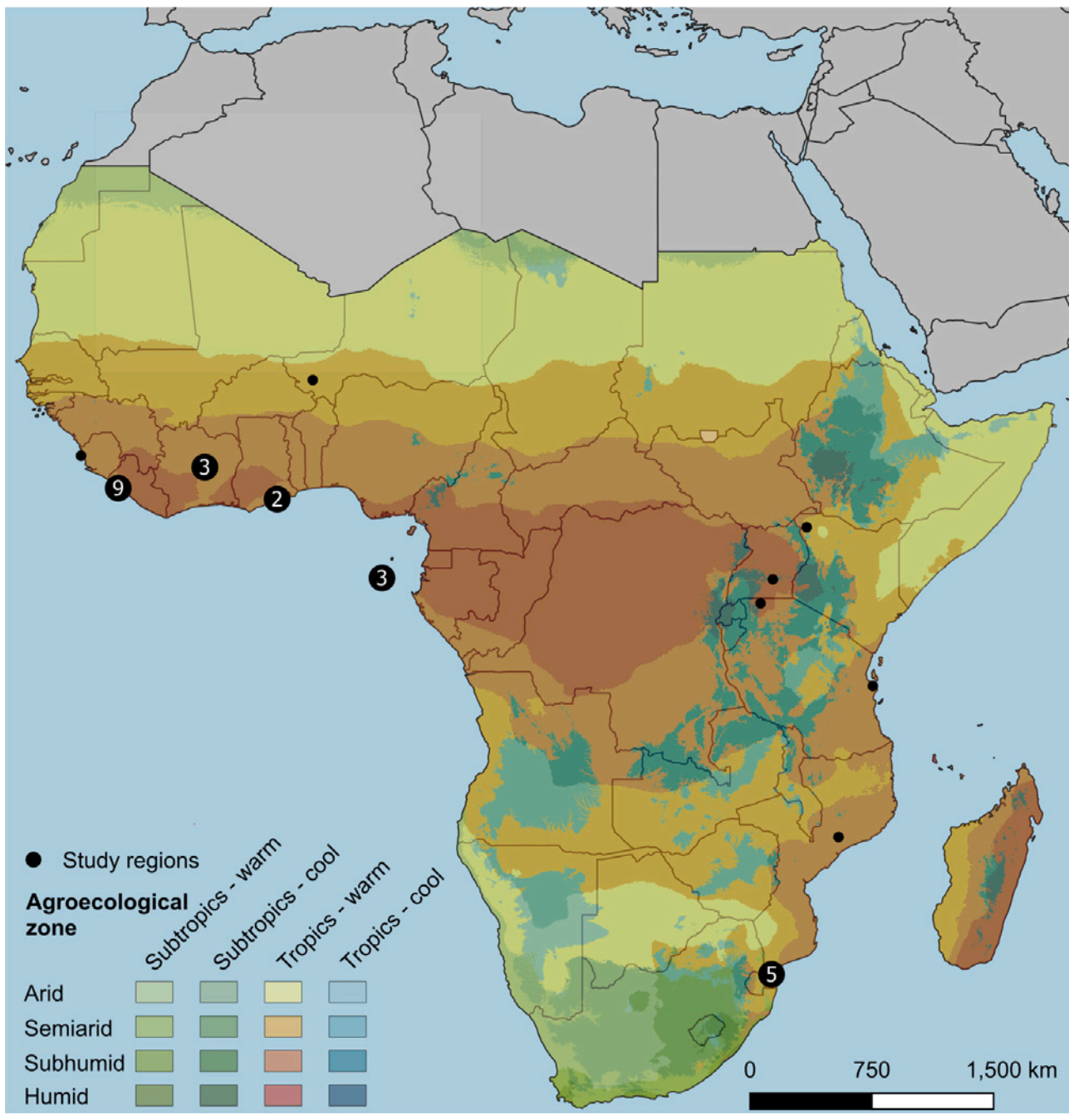

**Fig. A.1.** Geographic distribution of study regions across Sub-Saharan Africa. Map showing the location of UAV study regions (black circles) used to train and evaluate the open-access ODD-MSW detection model. Study sites span 10 countries across West, East and Southern Africa, encompassing diverse local practices and environmental contexts. The background map depicts agro-ecological zones (arid, semi-arid, sub-humid and humid) differentiated by tropical and subtropical thermal regimes, illustrating the broad environmental gradient represented in the UAV imagery (Sebastian, 2013). By sampling across heterogeneous environmental and agro-ecological contexts, the study design supports generalization of the open-access model for ODD-MSW detection.

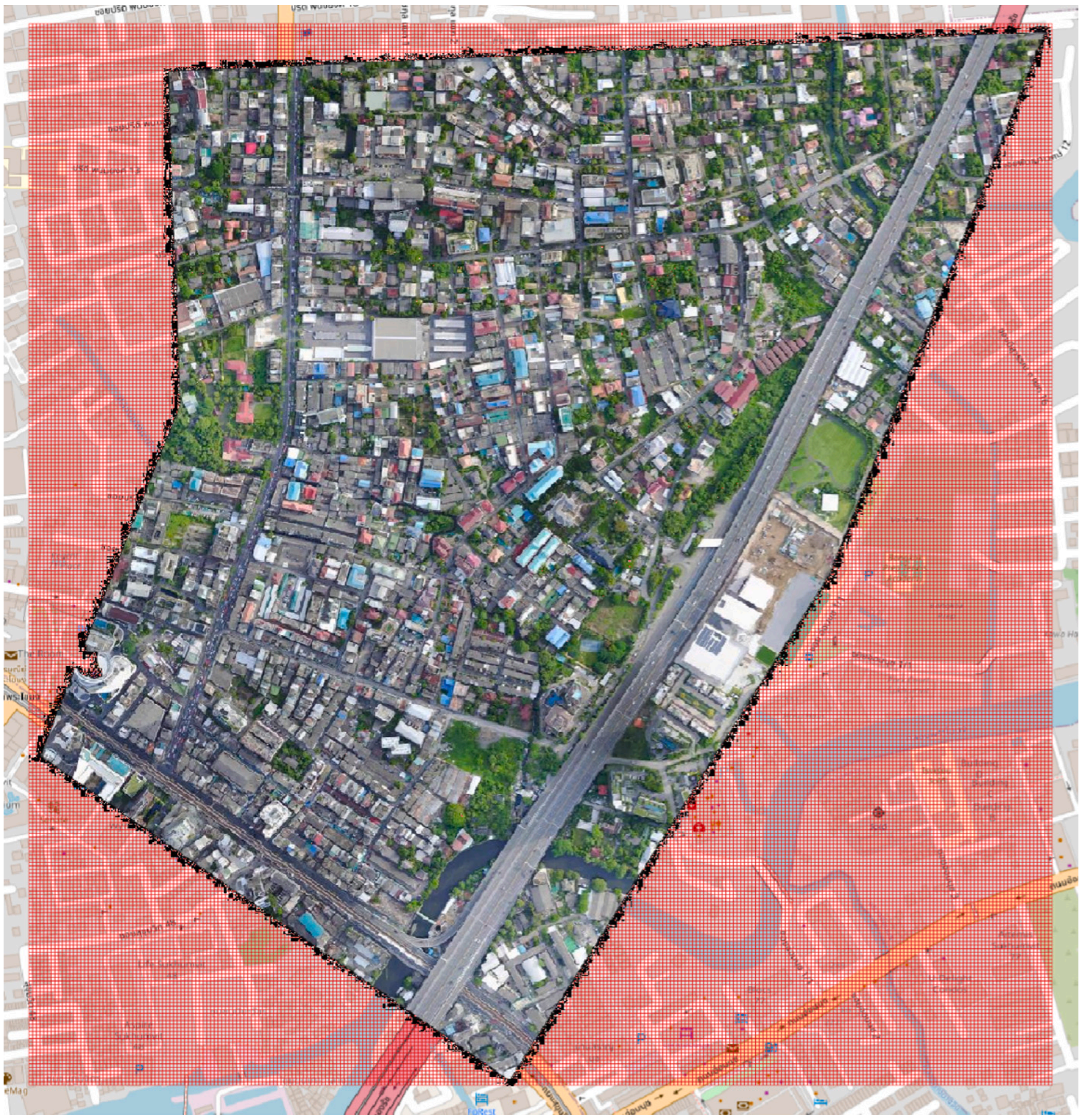

**Fig. A.2.** Example of a 5 × 5 m grid used for manual labeling of ODD-MSW, shown for a study site in Bangkok.

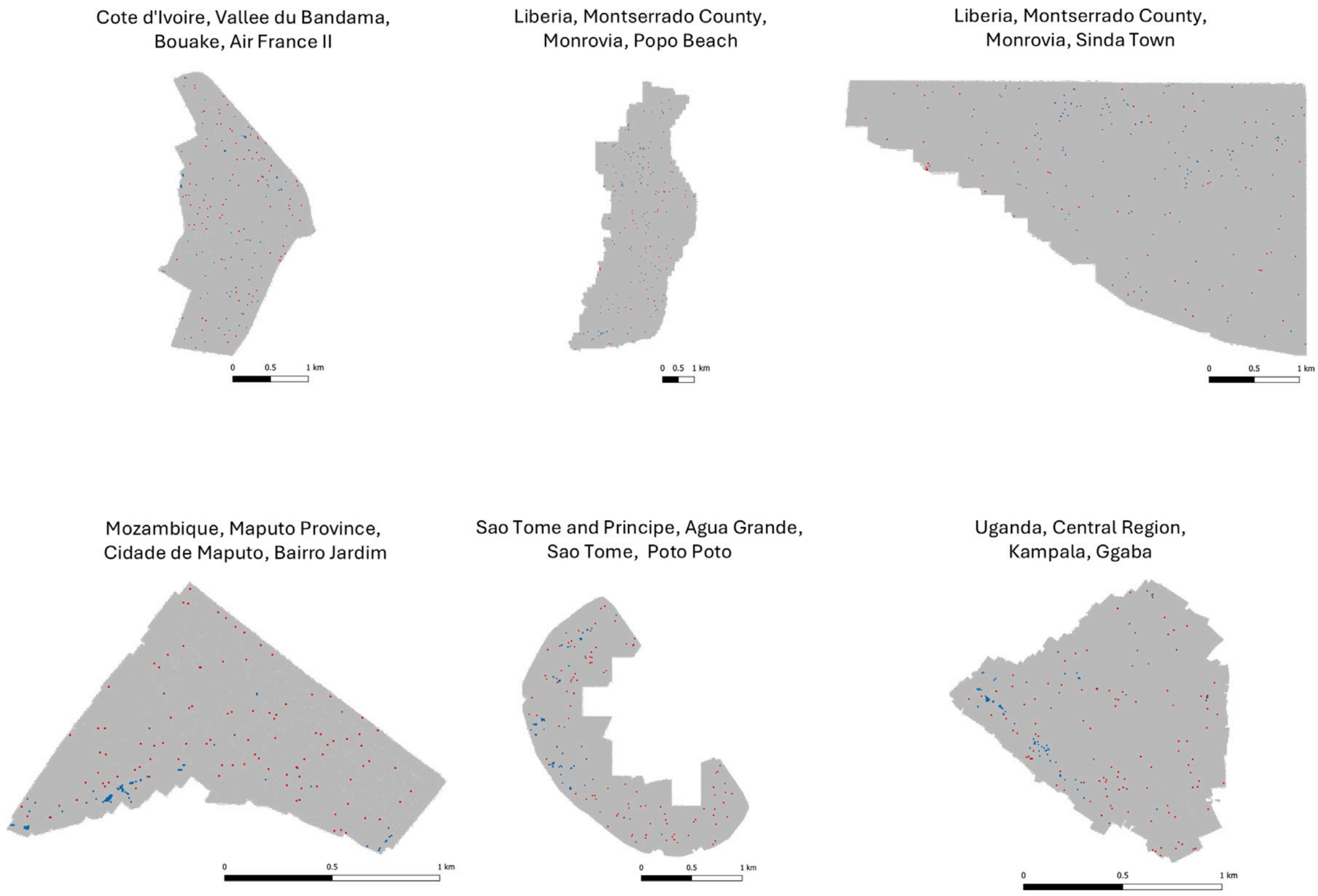

**Fig. A.3.** Representative regional labeling across study sites. Six of the 29 study areas are shown, with manual annotations applied at the regional scale. ODD-MSW-labeled tiles (red) and background-labeled tiles (blue) are distributed across each area, illustrating the spatial coverage of the annotation process. Each region contains 100 labeled tiles per class, providing balanced training data for deep learning-based classification.

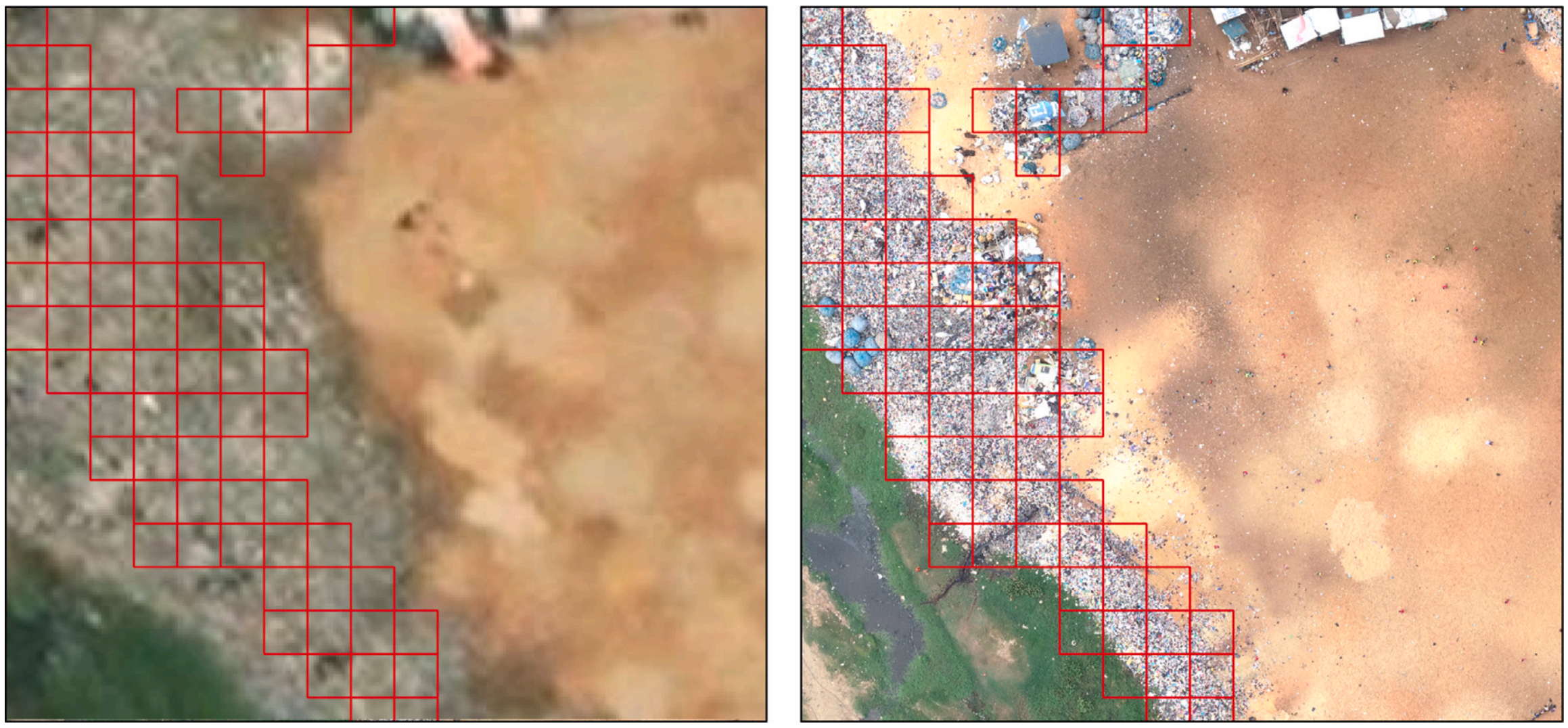

**Fig. A.4.** Side-by-side comparison of open-source satellite imagery and UAV imagery. The panels directly contrast spatial resolution and the capability to detect ODD-MSW, highlighting the greater detail and improved feature discernibility provided by UAV data relative to openly available satellite imagery.

**Table A.1**
Downloaded OpenAerialMap datasets used for model training and evaluation. Shown are dataset-level details, including country, state or region, city, place name, approximate scene-centroid coordinates, ground sampling distance (GSD), coverage area, OpenAerialMap (OAM) identifier, acquisition date and data provider. Every row was verified individually against the corresponding OpenAerialMap record. Acquisition dates are given in ISO 8601 format (YYYY-MM-DD) to avoid ambiguity in the ordering of day and month. Coordinates are given in decimal degrees (WGS 84, EPSG:4326), with positive values north and east.

| Country | State/Region | City | Place name | Centroid (lat, lon) | GSD | Coverage | OAM ID | Acquisition (YYYY-MM-DD) | Provider/Owner |
|---|---|---|---|---|---|---|---|---|---|
| Cote d'Ivoire | Vallee du Bandama | Bouake | Belleville I | 7.6820, −5.0430 | 5.00 cm | 13.66 km$^2$ | 666e4ea2f1cf8e0001fb2f64 | 2020-05-19 | IRD - MIVEGEC |
| | | | Air France II | 7.6905, −5.0135 | 5.00 cm | 7.94 km$^2$ | 666ef3bdf1cf8e0001fb2f6e | 2020-05-08 | IRD - MIVEGEC |
| | | | Koko | 7.7010, −5.0290 | 5.00 cm | 2.61 km$^2$ | 666b9b63f1cf8e0001fb2f16 | 2020-05-06 | IRD - MIVEGEC |
| Ghana | Greater Accra | Accra | Alajo | 5.5920, −0.2085 | 3.60 cm | 12.29 km$^2$ | 5be9bf8ac6c3bf0005896106 | 2018-11-12 | Makoko |
| | | | Old Fadama | 5.5480, −0.2205 | 5.00 cm | 2.08 km$^2$ | 66e40ff3cd0baa0001b6212f | 2024-08-27 | OpenStreetMap Ghana |
| Kenya | Turkana County | Kakuma | Kakuma | 3.7130, 34.8560 | 4.50 cm | 3.95 km$^2$ | 63b461953fb8c100063c5600 | 2022-11-22 | ESA Hub |
| Liberia | Montserrado County | Monrovia | Congo Town | 6.2865, −10.7565 | 5.10 cm | 4.41 km$^2$ | 65c4f779499b4d000186ee78 | 2020-01-19 | Humanitarian OpenStreetMap Team (HOT) |
| | | | Popo Beach | 6.3390, −10.8020 | 4.99 cm | 46.25 km$^2$ | 5dee77e79c3b1700059a3593 | 2019-10-02 | Uhurulabs |
| | | | Westpoint | 6.3345, −10.8095 | 5.31 cm | 13.91 km$^2$ | 65c4f080499b4d000186ee76 | 2020-01-19 | Humanitarian OpenStreetMap Team (HOT) |
| | | | Fluhaba | 6.2480, −10.6720 | 5.00 cm | 29.84 km$^2$ | 5e0ef7c515d478000501ea61 | 2019-12-09 | Uhurulabs |
| | | | Sinda Town | 6.2300, −10.5900 | 5.10 cm | 77.47 km$^2$ | 64d4eaba19cb3a000147a604 | 2023-08-09 | Humanitarian OpenStreetMap Team (HOT) |
| | | | 12th Street | 6.3010, −10.7960 | 5.08 cm | 59.81 km$^2$ | 5e557ce2642d040007b7c56f | 2020-02-23 | Uhurulabs |
| | | | SKD Boulevard | 6.2735, −10.7420 | 5.16 cm | 7.57 km$^2$ | 65c4f03f499b4d000186ee75 | 2020-01-19 | Humanitarian OpenStreetMap Team (HOT) |
| | | | Dixville | 6.3260, −10.7080 | 5.04 cm | 57.48 km$^2$ | 5e6a8cdb5abd57000732847c | 2020-02-23 | Uhurulabs |
| | | | Sinkor | 6.2930, −10.7830 | 5.06 cm | 15.75 km$^2$ | 65c4f27d499b4d000186ee77 | 2020-01-19 | Humanitarian OpenStreetMap Team (HOT) |
| Mozambique | Maputo Province | Cidade de Maputo | Bairro Jardim | −25.9095, 32.5760 | 5.00 cm | 2.60 km$^2$ | 5a7e18425a9ef7cb5d4efd59 | 2018-02-09 | Paolo Paron |
| | | | Chamanculo | −25.9490, 32.5710 | 3.55 cm | 3.24 km$^2$ | 64c3a7f13473010001ab8c4a | 2023-07-25 | #MapeandoMeuBairro |
| | | | Mafala | −25.9420, 32.5680 | 4.43 cm | 35.00 km$^2$ | 64a486bf64adbc00012e082e | 2023-07-24 | #MapeandoMeuBairro |
| | | | Maxaquene | −25.9345, 32.5835 | 4.43 cm | 40.00 km$^2$ | 66156a9fe89cf30001e0c3bb | 2024-04-08 | #MapeandoMeuBairro |
| | | | Bairro Matola | −25.9620, 32.4585 | 4.49 cm | 8.58 km$^2$ | 67421e365f60b100016fc070 | 2024-11-19 | #MapeandoMeuBairro |
| | Zambezia | Mocuba | Bairro Sacras | −16.8390, 36.9865 | 3.92 cm | 3.23 km$^2$ | 5dee39919c3b1700059a3586 | 2019-10-23 | INGC |
| Niger | Niamey Region | Niamey | Saga Sahara | 13.4810, 2.1490 | 5.89 cm | 1.94 km$^2$ | 5a25ae87bac48e5b1c51946f | 2017-11-15 | Drone Africa Service |
| Sao Tome and Principe | Agua Grande | Sao Tome | Poto Poto | 0.3390, 6.7340 | 5.92 cm | 5.82 km$^2$ | 59e62b943d6412ef7220a2a5 | 2017-02-28 | Drones Adventures |
| | | | Praia Luchinga | 0.3255, 6.7290 | 4.40 cm | 2.86 km$^2$ | 59e62b943d6412ef7220a28f | 2017-03-03 | Drones Adventures |
| | Me-Zochi | Bombom | Ganda II | 0.3050, 6.6890 | 4.53 cm | 1.30 km$^2$ | 59e62b943d6412ef7220a29d | 2017-02-28 | Drones Adventures |
| Sierra Leone | Western Area | Freetown | Freetown | 8.4790, −13.2340 | 3.95 cm | 50.00 km$^2$ | 69075f1de47603686de24fe8 | 2025-10-22 | DroneTM |
| Tanzania | Dar es Salaam | Dar es Salaam | Msimbazi | −6.8135, 39.2565 | 5.04 cm | 14.00 km$^2$ | 6a7c3a93a45360fc942a70ec | 2025-09-23 | OpenMapDevelopment Tanzania |
| | Kagera Region | Bukoba | Bukoba City | −1.3320, 31.8120 | 3.52 cm | 3.95 km$^2$ | 59e62b8a3d6412ef72209d69 | 2016-10-09 | WeRobotics |
| Uganda | Central Region | Kampala | Ggaba | 0.2745, 32.6395 | 3.53 cm | 2.02 km$^2$ | 5bc0ae9fc7e1cf0008e45e1f | 2018-09-17 | GeoGecko |

**Table A.2**
Site-specific model evaluation metrics across heterogeneous study locations. Performance metrics — including true positives (TP), false positives (FP), false negatives (FN), true negatives (TN), accuracy, precision, recall, and F1-score — are reported for the classification of ODD-MSW against background tiles across 29 distinct geographical locations. Place names correspond to those listed in Table A.1.

| Country | State/Region | City | Region name | TP | FP | FN | TN | Accuracy | Precision | Recall | F1 |
|---|---|---|---|---|---|---|---|---|---|---|---|
| Côte d'Ivoire | Vallée du Bandama | Bouáké | Belleville I | 12 | 2 | 4 | 12 | 0.8000 | 0.8571 | 0.7500 | 0.8000 |
| | | | Air France II | 12 | 1 | 2 | 15 | 0.9000 | 0.9231 | 0.8571 | 0.8889 |
| | | | Koko | 12 | 0 | 3 | 15 | 0.9000 | 1.0000 | 0.8000 | 0.8889 |
| Ghana | Greater Accra | Accra | Alajo | 14 | 0 | 2 | 14 | 0.9333 | 1.0000 | 0.8750 | 0.9333 |
| | | | Old Fadama | 14 | 1 | 1 | 14 | 0.9333 | 0.9333 | 0.9333 | 0.9333 |
| Kenya | Turkana County | Kakuma | Kakuma | 9 | 1 | 4 | 16 | 0.8333 | 0.9000 | 0.6923 | 0.7826 |
| Liberia | Montserrado County | Monrovia | Congo Town | 15 | 0 | 0 | 15 | 1.0000 | 1.0000 | 1.0000 | 1.0000 |
| | | | Popo Beach | 15 | 0 | 0 | 15 | 1.0000 | 1.0000 | 1.0000 | 1.0000 |
| | | | Westpoint | 15 | 3 | 0 | 12 | 0.9000 | 0.8333 | 1.0000 | 0.9091 |
| | | | Fluhaba | 12 | 1 | 2 | 15 | 0.9000 | 0.9231 | 0.8571 | 0.8889 |
| | | | Sinda Town | 13 | 0 | 1 | 16 | 0.9667 | 1.0000 | 0.9286 | 0.9630 |
| | | | 12th Street | 14 | 1 | 1 | 14 | 0.9333 | 0.9333 | 0.9333 | 0.9333 |
| | | | SKD Boulevard | 14 | 0 | 1 | 15 | 0.9667 | 1.0000 | 0.9333 | 0.9655 |
| | | | Dixville | 14 | 0 | 2 | 14 | 0.9333 | 1.0000 | 0.8750 | 0.9333 |
| | | | Sinkor | 14 | 1 | 1 | 14 | 0.9333 | 0.9333 | 0.9333 | 0.9333 |
| Mozambique | Maputo Province | Cidade de Maputo | Bairro Jardim | 14 | 1 | 1 | 14 | 0.9333 | 0.9333 | 0.9333 | 0.9333 |
| | | | Chamanculo | 13 | 0 | 2 | 15 | 0.9333 | 1.0000 | 0.8667 | 0.9286 |
| | | | Mafala | 15 | 0 | 0 | 15 | 1.0000 | 1.0000 | 1.0000 | 1.0000 |
| | | | Maxaquene | 15 | 1 | 2 | 12 | 0.9000 | 0.9375 | 0.8824 | 0.9091 |
| | | | Bairro Matola | 14 | 0 | 1 | 15 | 0.9667 | 1.0000 | 0.9333 | 0.9655 |
| | Zambézia | Mocuba | Bairro Sacras | 15 | 0 | 0 | 15 | 1.0000 | 1.0000 | 1.0000 | 1.0000 |
| Niger | Niamey Region | Niamey | Saga Sahara | 15 | 0 | 0 | 15 | 1.0000 | 1.0000 | 1.0000 | 1.0000 |
| São Tomé e Príncipe | Água Grande | São Tomé | Poto Poto | 14 | 1 | 1 | 14 | 0.9333 | 0.9333 | 0.9333 | 0.9333 |
| | | | Praia Luchinga | 14 | 2 | 1 | 13 | 0.9000 | 0.8750 | 0.9333 | 0.9032 |
| | Mé-Zóchi | Bombom | Ganda II | 14 | 0 | 1 | 15 | 0.9667 | 1.0000 | 0.9333 | 0.9655 |
| Sierra Leone | Western Area | Freetown | Freetown | 13 | 0 | 4 | 13 | 0.8667 | 1.0000 | 0.7647 | 0.8667 |
| Tanzania | Dar es Salaam Region | Dar es Salaam | Msimbazi | 14 | 1 | 1 | 14 | 0.9333 | 0.9333 | 0.9333 | 0.9333 |
| | Kagera Region | Bukoba | Bukoba City | 15 | 1 | 3 | 11 | 0.8667 | 0.9375 | 0.8333 | 0.8824 |
| Uganda | Central Region | Kampala | Ggaba | 13 | 0 | 3 | 14 | 0.9000 | 1.0000 | 0.8125 | 0.8966 |

**Table A.3**
Independent OpenAerialMap scenes used for the zero-shot transferability assessment. None of these scenes was used at any stage of model development. Reported are the continental region, country, city, OpenAerialMap (OAM) identifier, the number of independently annotated reference labels per class (background and ODD-MSW), the total number of labels, and the resulting ODD-MSW F1 score obtained with the frozen model. Annotation followed the labeling protocol of Section 2.2 without modification, and no fine-tuning, threshold re-calibration or region-specific adaptation was applied.

| Region | Country | City | OAM ID | Background labels | ODD-MSW labels | Total labels | F1 |
|---|---|---|---|---|---|---|---|
| Africa | Gambia | Kanifing | 67486e71e059320001709cb5 | 156 | 264 | 420 | 0.941 |
| | Malawi | Blantyre | 644f84655874aa0006657733 | 87 | 47 | 134 | 0.902 |
| Asia | Bangladesh | Chattogram | 6a1eb95693c47b7abc1c2270 | 104 | 104 | 208 | 0.913 |
| | Bangladesh | Dhaka | 652a6a10db601900014daad6 | 87 | 72 | 159 | 0.888 |
| | India | Vijayawada | 680883fecd738b2388b1efdf | 125 | 109 | 234 | 0.896 |
| | Iraq | Al Qurna | 60c46a01338cc80005cfdd17 | 117 | 104 | 221 | 0.874 |
| | Iraq | Al Qurna | 60c05ff4ace22e0007396bbd | 175 | 148 | 323 | 0.885 |
| | Mongolia | Ulaanbaatar | 59e62b7a3d6412ef7220952e | 126 | 99 | 225 | 0.831 |
| | Nepal | Changunarayan | 6a4b881d6332d62593123314 | 136 | 11 | 147 | 0.786 |
| | Philippines | Quezon City | 67f0c85c5d8e66413058cf72 | 36 | 13 | 49 | 0.812 |
| | Syria | Douma | 695780da7a6cf4cc4f7e0e13 | 136 | 20 | 156 | 0.769 |
| | Thailand | Bangkok | 59e62b963d6412ef7220a3bb | 120 | 16 | 136 | 0.824 |
| Europe | Turkey | Izmir | 66c4e8b90378ba0001bb5e0c | 99 | 9 | 108 | 0.742 |
| | Ukraine | Lviv | 6066d12ad8a0ef00061b9508 | 133 | 9 | 142 | 0.718 |
| | Ukraine | Lviv | 60c07dc9ace22e0007396bc1 | 152 | 80 | 232 | 0.857 |
| North America | Haiti | Les Cayes | 61266ef1b7012200056b5cbb | 233 | 234 | 467 | 0.968 |
| | Mexico | Querétaro | 68364d73000ce578cde9b9a5 | 106 | 52 | 158 | 0.879 |
| South America | Argentina | Pergamino | 62ea9f3c140bec00067db728 | 141 | 97 | 238 | 0.861 |
| | Argentina | Pergamino | 62eaa2f8140bec00067db72a | 153 | 94 | 247 | 0.873 |
| | Bolivia | La Paz | 67ef3bd5ca6f76f97a4b02e7 | 108 | 32 | 140 | 0.845 |
| | Bolivia | Oruro | 69bf20f6f9ab3980e7c29149 | 56 | 21 | 77 | 0.803 |
| | Brazil | Belo Horizonte | 6783c2c535bd090001afd100 | 45 | 13 | 58 | 0.797 |
| | Brazil | Belo Horizonte | 6783c37435bd090001afd102 | 111 | 18 | 129 | 0.836 |
| | Colombia | Medellín | 69efa9837c44e259af1ec5c4 | 109 | 38 | 147 | 0.892 |
| | Peru | Lima | 67dce2d9d1f1285c00fcb196 | 169 | 70 | 239 | 0.907 |
| **Total** | **19 countries** | **21 cities** | | **3020** | **1774** | **4794** | **0.867** |

## Data availability

The fine-tuned model weights, the inference and tiling code, the manual annotations and the derived tile-level predictions for all study regions are openly released. Code and model weights are available at Zenodo (https://doi.org/10.5281/zenodo.19731126). All input UAV imagery is third-party open data obtained from OpenAerialMap (https://openaerialmap.org) and is fully identified by the OpenAerialMap identifiers listed in Table A.1 and Table A.3, so every scene used here can be retrieved independently. The SHDI is available from the Global Data Lab (Smits & Permanyer, 2019), gridded population estimates from WorldPop (2026), and the infrastructure deficit index from Bettencourt and Marchio (2025). Model weights and code are released under an open licence permitting reuse and modification, and the released model is the exact artifact evaluated in Sections 3.1–3.3, rather than a subsequently retrained version.